\documentclass[journal]{IEEEtran}
\usepackage{times}  
\usepackage{helvet}  
\usepackage{courier}  
\usepackage[hyphens]{url}  
\usepackage{graphicx} 
\usepackage{caption} 
\usepackage{algorithm}
\usepackage{algorithmic}
\usepackage{multirow}
\usepackage{amsmath}
\usepackage{colortbl}  
\usepackage{xcolor}
\usepackage{amsfonts}       
\usepackage{newfloat}
\usepackage{listings}
\usepackage{booktabs}
\usepackage[colorlinks,
linkcolor=red,
anchorcolor=red,
citecolor=blue]{hyperref}
\usepackage[switch]{lineno}

\definecolor{baselinecolor}{gray}{.9}

\newcommand{\revise}[1]{\textcolor{black}{#1}}
\newcommand{\newrevise}[1]{\textcolor{black}{#1}}
\newcommand{\minorrevise}[1]{\textcolor{black}{#1}}
\DeclareCaptionStyle{ruled}{labelfont=normalfont,labelsep=colon,strut=off} 
\floatstyle{ruled}
\newfloat{listing}{tb}{lst}{}
\floatname{listing}{Listing}
\title{Multi-Condition Latent Diffusion Network for \\Scene-Aware Neural Human Motion Prediction}
\author{
	Xuehao Gao, \textit{Student Member, IEEE,}
	Yang Yang, \textit{Member, IEEE,}
	Yang Wu, \textit{Member, IEEE,} \\
	Shaoyi Du, \textit{Member, IEEE,}
	and	Guo-Jun Qi, \textit{Fellow, IEEE} 
	\thanks{This work was supported by the National Key Research and Development Program of China under Grant No.2018AAA0102500. (Corresponding author: Yang Yang.)}
	\thanks{Xuehao Gao, Shaoyi Du, and Yang Yang are with Xi’an Jiaotong University, Xi’an 710049, China (e-mail: gaoxuehao.xjtu@gmail.com; dushaoyi@gmail.com; yyang@mail.xjtu.edu.cn).}
	\thanks{Yang Wu is with Tencent AI Lab. Shenzhen 518057, China (e-mail: dylanywu@tencent.com).}
	\thanks{Guo-Jun Qi is with the School of Engineering, Westlake University, Hangzhou 310030, China, and also with the OPPO U.S. Research Center, Bellevue, WA 98004 USA (e-mail: guojunq@gmail.com).}
}

\begin{document}
	
	\maketitle
	
	\begin{abstract}
		Inferring 3D human motion is fundamental in many applications, including understanding human activity and analyzing one's intention. While many fruitful efforts have been made to human motion prediction, most approaches focus on pose-driven prediction and inferring human motion in isolation from the contextual environment, thus leaving the body location movement in the scene behind. However, real-world human movements are goal-directed and highly influenced by the spatial layout of their surrounding scenes. In this paper, instead of planning future human motion in a “dark” room, we propose a Multi-Condition Latent Diffusion network (MCLD) that reformulates the human motion prediction task as a multi-condition joint inference problem based on the given historical 3D body motion and the current 3D scene contexts. \revise{Specifically, instead of directly modeling joint distribution over the raw motion sequences, MCLD performs a conditional diffusion process within the latent embedding space, characterizing the cross-modal mapping from the past body movement and current scene context condition embeddings to the future human motion embedding.} Extensive experiments on large-scale human motion prediction datasets demonstrate that our MCLD achieves significant improvements over the state-of-the-art methods on both realistic and diverse predictions.

	\end{abstract}
	
	\begin{IEEEkeywords}
		3D human motion prediction, latent diffusion model, multi-condition inference.
	\end{IEEEkeywords}
	
	\section{Introduction}
	\label{intro}
	\IEEEPARstart{H}{uman} motion prediction forecasts future poses and trajectories from a few time steps of historical human motion. Motion prediction helps intelligent machines to understand human intention by analyzing their behaviors and then plan own reactions \cite{DBLP:cvpr/xuehao, DBLP:journals/pr/YangMZWGC22, lyu20223d, DBLP:journals/tip/LiTZFL21}. These predictive techniques are essential in many real-world applications, including human-machine interaction \cite{DBLP:journals/pami/KoppulaS16,DBLP:conf/icml/KoppulaS13}, autonomous driving \cite{DBLP:journals/ral/LefkopoulosMDZ21,DBLP:journals/spm/FernandoDSF21} and surveillance systems \cite{DBLP:conf/mmasia/TangZY0Y20}. Although recent human motion prediction works have made great efforts and fruitful progress \cite{DBLP:journals/ijcv/MaoLSL21,DBLP:conf/iccv/MaoLS21,DBLP:conf/iccv/LiuS0SCH021,DBLP:conf/cvpr/CuiS21,DBLP:conf/aaai/LiuLWCHJ21}, most of them focus on pose-driven motion prediction and inferring human motion in isolation from the surrounding environment, thus leaving the body location movement in the scene behind \cite{DBLP:conf/iccv/AliakbarianSPGS21,DBLP:journals/ral/ZhuCBA21,DBLP:journals/tip/LiCZZWT21,DBLP:conf/iros/PostnikovGF21}. The key insight missing is that human pose and trajectory are goal-directed and closely influenced by the spatial layout of their surrounding environments, suggesting a contextual human-scene interaction can be further introduced for improving human motion prediction. We can imagine a scenario where a person walks around furniture and then goes down the stairs. In this case, the geometry structure of a 3D scene enforces multiple context constraints on the future human motion ($e.g.$, two legs avoiding collision with the furniture and feet touching the steps). Therefore, instead of planning future human motion in a “dark” room, we infer indoor future human motions conditioned on past body movements and 3D current scene contexts, exploring the probabilistic mapping behind scene-aware human motion prediction task.
	
     However, how to infer one's future motion from his past body movements and current scene context remains under-explored. The core challenges behind the scene-aware human motion prediction task are: \textbf{\textit{(1) \newrevise{Joint inference based on multiple conditions.}}} The conditional dependencies of future human motions on past body movements and current scene contexts are intertwined. \newrevise{Therefore, as a challenging joint inference problem, scene-aware human motion prediction focuses on predicting a realistic human motion sequence that is jointly compatible with both given historical body motions and current scene contexts.} \textbf{\textit{(2) Many-to-many mapping between given conditions and future human motions.}} The non-deterministic many-to-many problem behind the scene-aware human motion prediction task makes synthesizing realistic and diverse 3D human motions much more challenging. \newrevise{For example, diverse human motion samples with different poses and trajectories may both conform to an intended prediction of the same given motion-scene condition contexts.} \newrevise{Therefore, suffering from these two challenges, an ideal human motion prediction system should bridge a probabilistic causality mapping between an indoor human motion sequence and its joint conditions of past body movements and current scene context.}

	\begin{figure*}[!t]
		\centering
		\includegraphics[width=1\textwidth]{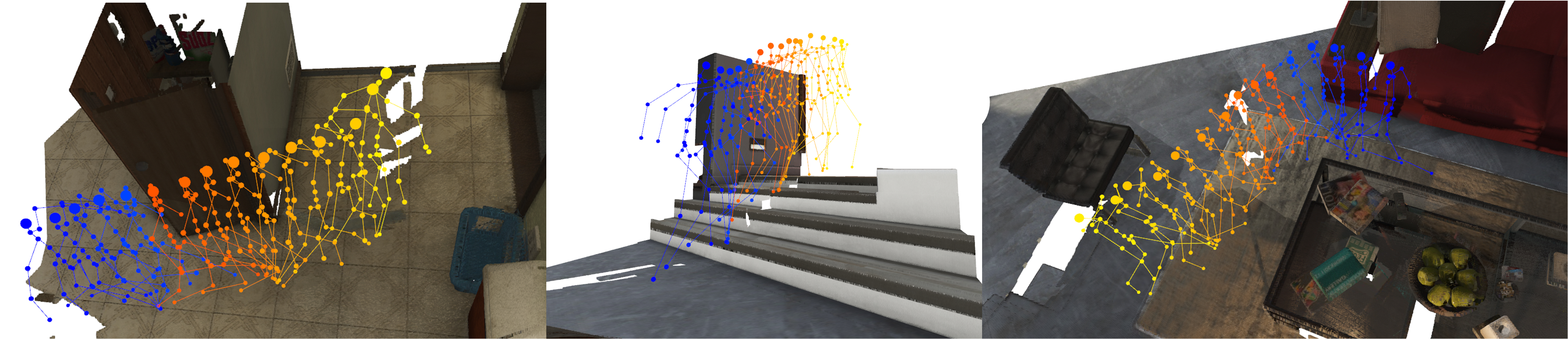}
		\caption{Three examples of our predicted long-term 3D human motion in the 3D scene. Given the past motion history (blue skeletons), we forecast realistic future pose (red skeletons) with diverse human-scene interactions. The color of the predicted pose is changed over frames.}
		\label{fig1}
	\end{figure*}

    Among recent methods, Variational Autoencoders (VAEs) and Generative Adversarial Networks (GANs) are widely adopted for human motion prediction \cite{mao2019learning,mao2021generating, xu2022diverse, dang2022diverse,nikdel2023dmmgan, kundu2019bihmp, xu2023actformer}. \newrevise{However, VAEs limit their learned distributions to normal latent distributions, thus enforcing strong prior assumptions on target human motion distributions. Without an effective training strategy, GANs tend to suffer from mode collapse and vanishing gradient problems.}  \newrevise{Compared with these generation schemes, diffusion models are free from prior assumptions on the target human motion distribution, thus significantly facilitating probabilistic many-to-many mapping learning in the scene-aware human motion prediction task.} Notably, instead of directly modeling the joint distribution over the raw motion sequences, we develop a latent diffusion model that characterizes the conditional dependencies between motion history, scene context, and future body movements in their feature embedding spaces. Specifically, we first deploy a powerful VAE module on a future human motion sequence to extract its low-dimensional latent representation. Then, instead of using a diffusion model to establish direct connections between the raw motion sequences and their conditional inputs, we perform a diffusion process on the embedded motion latent space. Finally, in this paper, we propose a novel Multi-Condition Latent Diffusion model (MLCD) that could predict realistic and diverse future motion sequences conforming to the given multiple conditions of past human movements and current scene context.

    \revise{Besides a basic latent-based diffusion model, MCLD also proposes three core components to improve its realistic motion prediction: \textbf{\textit{(1) Key Region Proposal module for refining scene conditions.}} Considering that indoor scene objects hardly bring environmental constraints into body motion far away from them, MCLD thus proposes a Key Region Proposal module (KRP) to adaptively demarcate a localized interaction-related region from a given initial 3D scene for each human motion sample, significantly reducing the redundancy of the raw 3D scene input; \textbf{\textit{(2) Multi-Attention Encoder for extracting richer feature embeddings from condition inputs.}} Different dependencies within and between historical 3D body joints and current 3D scene points derive multiple feature embeddings, including body dynamics, scene geometries, and body-scene interactions. To model these multiple representations, MCLD develops a Multi-Attention Encoder (MAE) on condition inputs to extract richer feature embeddings from them; \textbf{\textit{(3) Multi-Condition Fusion for dynamically integrating extracted multiple condition features.}} The respective influences of 3D human motion history and 3D scene context on future human motion are sample-related. To this end, MCLD proposes a Multi-Condition Fusion module (MCF) to infer the adaptive response of extracted multiple condition features and dynamically integrate them at each input sample. As shown in Fig. \ref{fig1}, by coupling MLCD with these three effective components, we develop a powerful scene-aware human motion prediction system that significantly outperforms state-of-the-art methods on motion realism and diversity.}

    \revise{Our technical contributions are summarized in the following:} 
	\begin{itemize}
		\item \revise{We propose a novel multi-condition latent diffusion strategy (MCLD) that characterizes an effective probabilistic mapping from the joint conditions of past 3D body movements and current 3D scene context to the future human motion, improving scene-aware 3D human motion prediction on motion realism and diversity.} 
  
		\item \revise{We propose an effective key region proposal module (KRP) that dynamically demarcates a localized interaction-related region for each human motion sample, significantly benefiting a robust scene-aware human motion prediction against redundant scene noise.}
  
       \item \revise{We develop a multi-attention encoder (MAE) that acts as a powerful feature extractor to encode richer latent feature representations from given conditional inputs, improving realistic human motion prediction.}

        \item \revise{We propose a multi-condition fusion module (MCF) that infers the dynamic response of feature embeddings extracted from condition inputs and adaptively integrates them at each sample, significantly enlarging model adaptiveness.}   

\end{itemize}

	\begin{figure*}[!t]
	\centering
	\includegraphics[width=1\textwidth]{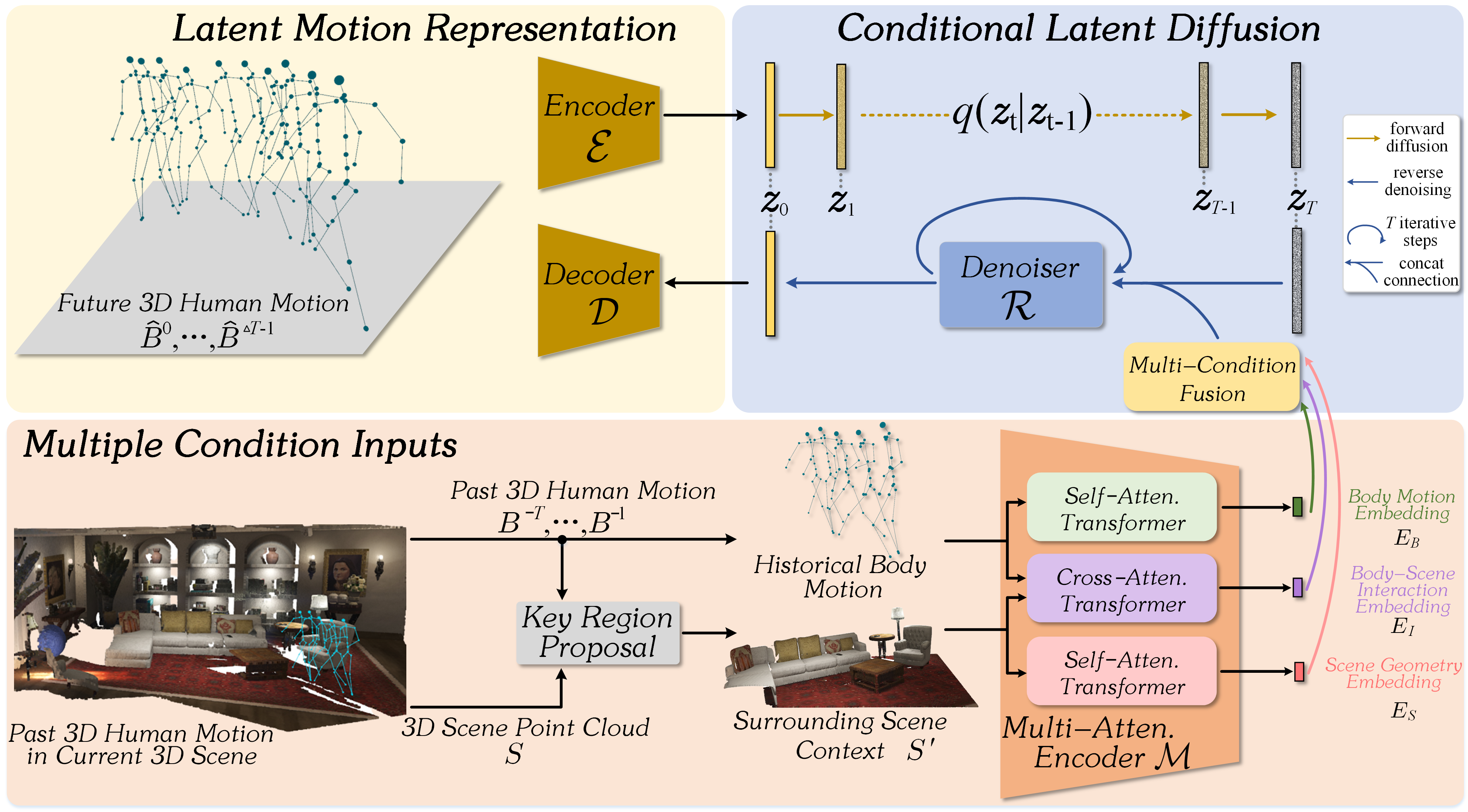}
	\caption{Architecture Overview. MCLD consists of a VAE model and a multi-condition latent-based diffusion model. MCLD proposes a two-stage training scheme: first adopting encoding-decoding reconstruction loss to optimize the VAE model and learn effective latent representation $z_{0}$ of future body movements ($i.e.$, $\widehat{B}^{0}, \cdots, \widehat{B}^{\Delta T -1}$); then adopting the noising-denoising strategy to optimize a latent conditional diffusion model and characterize the probabilistic mapping from the joint conditions of past body movements and current scene contexts to future human motion.}
	\label{pipeline}
\end{figure*}

	\section{Related Work}
	\subsection{Human Motion Prediction} \revise{Human motion task focuses on developing a powerful motion predictor that infers realistic future human motions from a given condition signal, such as past body poses, past movement trajectory, current scene context, \textit{etc.}.} Most human motion prediction methods are pose-driven and infer future body movements conditioned on a historical 3D skeleton-based human motion sequence \cite{DBLP:journals/natmi/PangZCTYL20,DBLP:conf/aaai/YanXL18,DBLP:conf/cvpr/LiuZC0O20,DBLP:conf/accv/001800L20,DBLP:journals/tip/WangDCF21,wang2022skeleton}. Since the topology of the human skeleton is a natural graph, many works adapt graph convolution networks (GCN) to extract spatial-temporal features \cite{DBLP:conf/iccv/DangNLZL21,DBLP:conf/iccv/SofianosSFG21,DBLP:journals/corr/abs-2203-01474,DBLP:conf/iccv/MaoLSL19,DBLP:conf/cvpr/CuiSY20,DBLP:journals/corr/abs-2112-15012,DBLP:conf/cvpr/LiCZZW020}. Specifically, most GCN-based approaches deploy interleaving spatial-only and temporal-only modules to extract intra-frame posture and inter-frame trajectory patterns, respectively \cite{DBLP:journals/tip/ZhuSLL23,DBLP:conf/preregister/Gao0D20,10081331,gao2021efficient}. In recent years, inspired by the effectiveness of the self-attention mechanism \cite{DBLP:conf/cvpr/0004GGH18,DBLP:conf/nips/VaswaniSPUJGKP17} in long-range modeling, many methods \cite{DBLP:conf/3dim/AksanKCH21,DBLP:journals/cviu/PlizzariCM21,DBLP:conf/mm/ZhangWLDG21} deploy transformer-based backbone to enlarge receptive fields and make distant neighbor nodes reachable. However, an ideal human motion prediction system should look beyond motion-specific conditions rather than leaving the human-scene interaction and scene constraints behind. We also would like to emphasize that we are not performing a human motion synthesis task since the end body location is agnostic in our predictive task. 
	
	\subsection{Human-Scene Interaction Modeling}
	Considering that real-world human motions are compatible with their surrounding scene contexts, future human motions are joint conditional distributions of body joints within the scene context \cite{starke2019neural,holden2016deep, wang2022humanise,li2021ai,wang2021synthesizing,huang2023diffusion}. Meanwhile, since daily human actions are goal-directed, the surrounding scene context is also crucial to analyze human intentions and predict his future poses and body movements. Therefore, besides human motion history, human-scene interaction inside the current scene context serves as another condition for future body motion prediction. In this case, instead of planning future human motion in isolation from the surrounding environment, scene-aware human motion prediction study has attracted more and more attention, gradually taking promising steps toward realistic motion generation. As an initial attempt, GPP-Net \cite{DBLP:conf/eccv/CaoGMCVM20} represents the scene with background images and constructs a convolutional network to implicitly encode human-scene interaction in the pixel space. Similarly, SocialPool \cite{DBLP:journals/ral/AdeliARNR20} deploys a GRU-based network to learn body dynamics from 3D skeletons and extract the scene context lying in RGB-based background videos. Skeleton-Graph \cite{DBLP:journals/corr/abs-2109-10257} develops GPP-Net by adopting graph convolutions to model body dynamics. However, these RGB-based appearance patterns are suffering from insufficient scene constraints and non-robust human-scene interaction modeling. Thus, how to effectively characterize scene context information for realistic human motion prediction is a fundamental yet under-explored task.

		\begin{figure*}[t]
		\centering
		\includegraphics[width=1\textwidth]{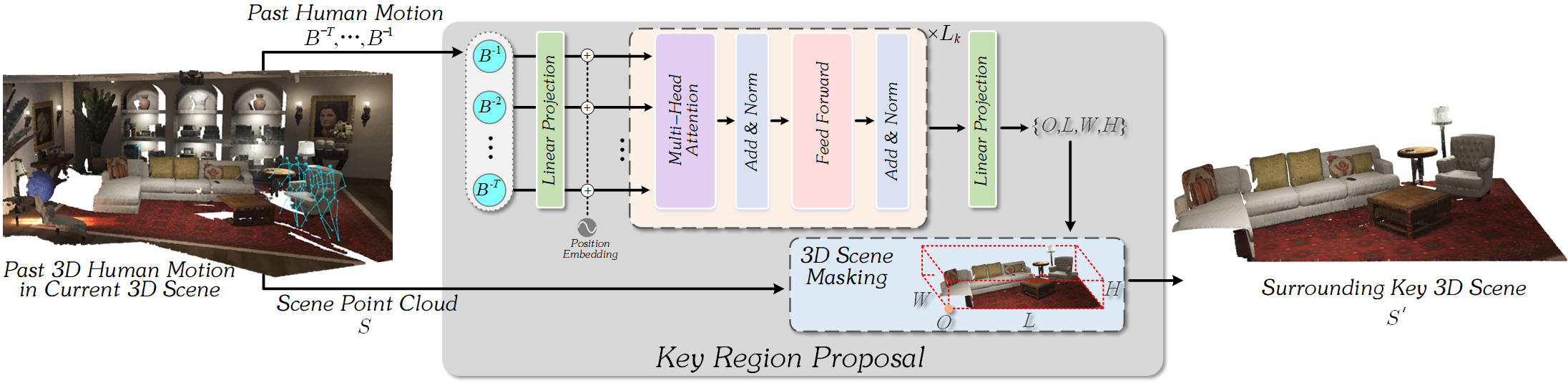}
		\caption{Key Region Proposal Module. Given a 3D scene point cloud ($i.e., S$) and a 3D human motion sequence ($i.e., B^{-}=\{B^{-T}, \dots, B^{-1}\}$) inside it, key region proposal module aims at adaptively demarcating a localized interaction-related 3D region $S^{\prime}$ from $S$ for each $B^{-}$ sample, significantly reducing the redundancy of the initial scene input $S$.}
		\label{krp_pipeline}
	\end{figure*}
 
	\subsection{Denoising Diffusion Probabilistic Models}
	\minorrevise{Different generation models are widely explored in the human motion prediction task, including VAEs \cite{ding2024expressive, xu2022stochastic, mahdavian2023stpotr, DBLP:conf/iccv/HassanCVSYZB21, DBLP:conf/eccv/CaoGMCVM20}, GANs \cite{zhao2023bidirectional, nikdel2023dmmgan, yasar2023vader, kundu2019bihmp}, and flow-based models \cite{zand2023flow, yuan2020dlow}. For example, EAI \cite{ding2024expressive} develops a VAE-based encode-decode framework to predict both coarse (body joints) and fine-grained (gestures) activities collaboratively. Actformer \cite{xu2023actformer} incorporates the Transformer-based motion generator into GAN, known for high-quality generative modeling. However, previous experiments indicate that these generative model both suffer from their potential limitations \cite{lyu20223d}. For example, VAEs limit their learned distributions to normal latent distributions, thus enforcing strong prior assumptions on target human motion distributions. Without an effective training strategy, GANs tend to suffer from mode collapse and vanishing gradient problems. As an emerging yet promising generative framework, diffusion model outperforms others in being free from prior assumptions on the target distribution.}
	
	Inspired by the stochastic diffusion process in Thermodynamics, diffusion models learn the distribution of a target sample by denoising its noised signal \cite{DBLP:conf/icml/Sohl-DicksteinW15, DBLP:conf/nips/0011E20, DBLP:journals/corr/abs-2209-00796}. Recently, diffusion and its variants are widely adopted by various tasks, including text-to-image synthesis \cite{saharia2022photorealistic,ramesh2022hierarchical,zhu2022label} text-to-video synthesis \cite{wang2023videofactory,blattmann2023align}, text-to-3D synthesis \cite{gao2024guess,tevet2022human}, and cross-modality recognition \cite{li2023graph,flaborea2023multimodal}. However, how to develop a powerful diffusion-based prediction system for inferring indoor 3D human motions still remains under-explored and confronts various challenges, such as joint inference based on motion-scene multi-condition contexts, robustness against scene condition redundancy, and body-scene interaction embedding. As an initial attempt, MCLD improve realistic indoor 3D human motion prediction with a series of powerful modules, including key region proposal, latent multi-condition diffusion, and multi-attention encoder. we hope MCLD will inspire more investigation in the community.

	
	\section{Problem Formulation}
	Scene-aware human prediction system aims at inferring realistic future body movements from the given body motion history and current scene context. Mathematically, let $S \in \mathbb{R}^{N_{s}\times3}$ be a 3D scene point cloud with $N_{s}$ points. $B^{-}=[B^{-T}, \dots, B^{-1}]$ be a $T$-frame 3D body motion history inside $S$, where $B^{t}\in \mathbb{R}^{N_{b}\times3}$ denotes the skeleton at $t$-th frame with $N_{b}$ body joints. In motion prediction, $B^{+} = [B^{0}, \dots, B^{\Delta T-1}]$ represents the future body skeletons at next $\Delta T$ frames. In this paper, we aim to develop a powerful predictor $\mathcal{F}_{\text {pred}}(\cdot)$ to infer the future motion $\widehat{B}^{+}=\mathcal{F}_{\text {pred}}(B^{-},S)$ to approximate its ground-truth $B^{+}$.

	\section{Methodology}
	\label{formulate}
	As shown in Fig. \ref{pipeline}, in this paper, we propose MCLD that performs a diffusion process in the latent space to learn the powerful cross-modal mapping from the past body movement and current scene context conditions to the future human motion. In the following, we describe the core components of MCLD and elaborate on their technical details.
	
	\subsection{Latent Motion Representation}
	\label{sec_lmr}
	In the training stage, we first deploy a transformer-based Variational AutoEncoder $\mathcal{V}$ on the future human motion $B^{+}$ to encode it into a low-dimensional latent space $\mathcal{Z}$ and learn its feature embedding $z$. Specifically, $\mathcal{V}$ consists of a $L_{v}$-layer transformer-based encoder $\mathcal{E}$ and decoder $\mathcal{D}$ as $\mathcal{V} = \{ \mathcal{E}, \mathcal{D} \}$. $\mathcal{E}$ uses the embedded distribution tokens as Gaussian distribution parameter $\mu$ and $\sigma$ of the latent motion space $\mathcal{Z}$ to reparameterize the extracted latent motion embedding $z$. Then, given $z$, the decoder $\mathcal{D}$ is encouraged to reconstruct the 3D motion sequence $\overline{B}^{+}$ that approximate its ground-truth $B^{+}$. With this encoding-decoding learning strategy, we extract the effective $C_{e}$-dimensional latent representation $z \in \mathbb{R}^{C_{e}}$ of the future human motion $B^{+}$. In the first training stage of MCLD, we optimize all VAE components ($i.e., \mathcal{E}, \mathcal{D}$) with two objectives: $\mathcal{L}_{mr}$ and $\mathcal{L}_{kl}$. Specifically, $\mathcal{L}_{mr}$ defines a motion reconstruction loss and focuses on learning an effective latent motion representation $z$. To regularize the latent space $\mathcal{Z}$, $\mathcal{L}_{kl}$ defines a Kullback-Leibler distance between $q\left(z \mid B^{+}\right) = \mathcal{N}\left(z ; \mathcal{E}_\mu, \mathcal{E}_{\sigma^2}\right)$ and a standard Gaussion distribution $\mathcal{N}(z ; 0,1)$. Finally, the loss function of the first training stage of MCLD is defined as:
	\begin{equation}
		\begin{aligned}
	\mathcal{L}_{\mathcal{V}} &= \lambda_{mr} \mathcal{L}_{mr} + \lambda_{kl} \mathcal{L}_{kl} \\
	           &= \lambda_{mr} \left\|B^{+}-\mathcal{D}\left(\mathcal{E}\left(B^{+}\right)\right)\right\|_2 \\
	           &+ \lambda_{kl} \textit{\textbf{KL}}\left(\mathcal{N}(\mu, \sigma^2) || \mathcal{N}(0, 1)\right), 
		\end{aligned}
		\label{eq.1}
	\end{equation}
	where $\lambda_{mr}$ and $\lambda_{kl}$ represents the weight of $\mathcal{L}_{mr}$ and $\mathcal{L}_{kl}$, respectively.    

 	\begin{figure*}[t]
		\centering
		\includegraphics[width=1\textwidth]{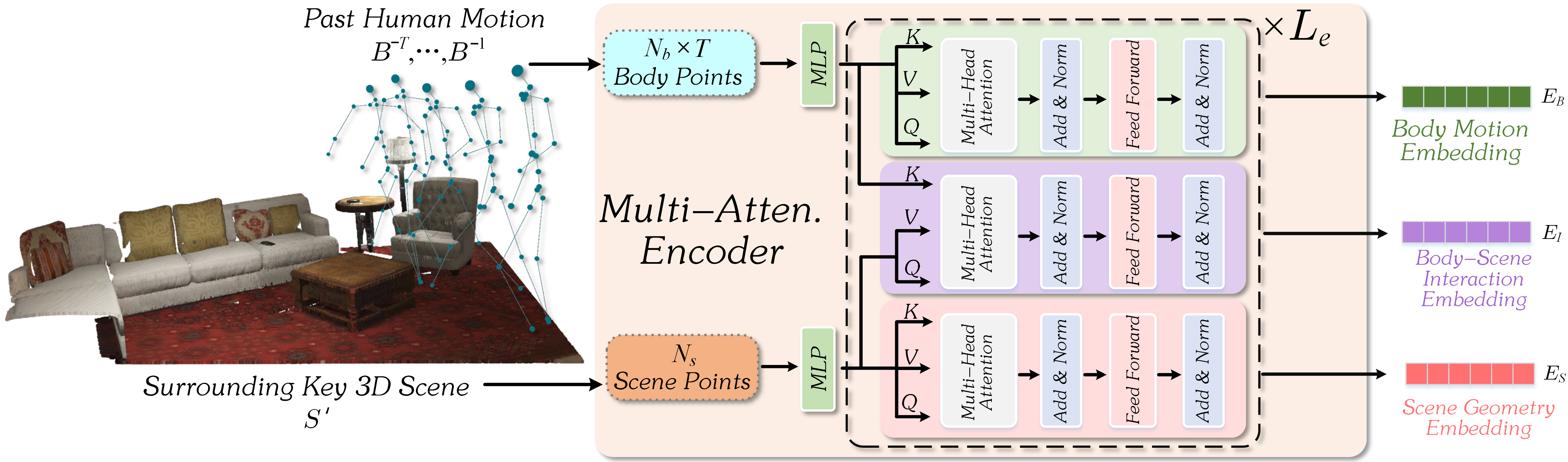}
		\caption{Multi-Attention Encoder Module. Considering there are multiple dependencies within and between body and scene points, including body motion, scene geometry, and body-scene interaction, we deploy a transformer-based $L_{e}$-layer multi-attention encoder on $S^{\prime}$ and $B^{-}$ to extract these latent embeddings jointly.}
		\label{mae_pipeline}
	\end{figure*}
	
	\subsection{Key Region Proposal}
	\label{key region proposal}
	Since scene objects hardly bring environmental constraints to body motions far away from them, we first develop a powerful key region proposal (KRP) module to dynamically demarcate a localized interaction-related region $S^{\prime}$ from $S$ for each human motion history sample $B^{-}$. By reducing the redundancy of the initial scene input $S$, KRP encourages a human motion prediction system to focus on the interactive context and isolate it from unnecessary scene noises. Specifically, different movement patterns of $B^{-}$ ($e.g.,$ action classes and motion trajectories) would deliver different cues for analyzing its key region. For example, the key region for \textit{sitting} tends to be narrower than for \textit{walking}; the key regions for \textit{walking left} and \textit{walking right} may be located in different directions. Therefore, we first develop a KRP module that takes the human motion history $B^{-}$ as input and infers the location and range of its 3D key scene region $S^{\prime}$. Specifically, as shown in Fig. \ref{krp_pipeline}, KRP first deploys a $L_{k}$-layer transformer with $h_{k}$ self-attention heads on $B^{-}$ to extract its body movement features. Then, based on the extracted body motion features, a linear projection with one hidden layer regresses location and range factors related to the key region proposal: origin position $O = \{O_{x}, O_{y}, O_{z}\}$, length $L$, width $W$, and height $H$. \revise{These parameters determine a 3D cube range $R$ inside $S$. Based on $R$, we define a binarized mask $M$ for the 3D scene $S$. Specifically, given an arbitrary point $S_{i}$ of scene point cloud $S$, its binarized mask value depends on whether $S_{i}$ is inside $R$: if $S_{i}$ is inside $R$, then $M_{i}=1$; otherwise $M_{i}=0$. Therefore, the 3D key scene region $S^{\prime}$ selected from $S$ is formulated as:}      

    \begin{equation}
		\begin{aligned}
	S^{\prime} &= S \odot M, \\
    \text{where} \; M_{i} & = \begin{cases} 1 & \text{ if } S_{i} \; \text{is inside} \; R, \\ 0 & \text { otherwise },\end{cases}
		\end{aligned}
  \label{mask}
	\end{equation}
 
 \revise{In Eq.\ref{mask}, $\odot$ denotes the element-wise product. Finally, based on the motion patterns of human motion sample $B^{-}$, the proposed KRP module adaptively infers its 3D key region $S^{\prime}$ inside $S$, significantly reducing the redundancy of original scene input $S$.} 

	\begin{figure*}[t]
		\centering
		\includegraphics[width=1\textwidth]{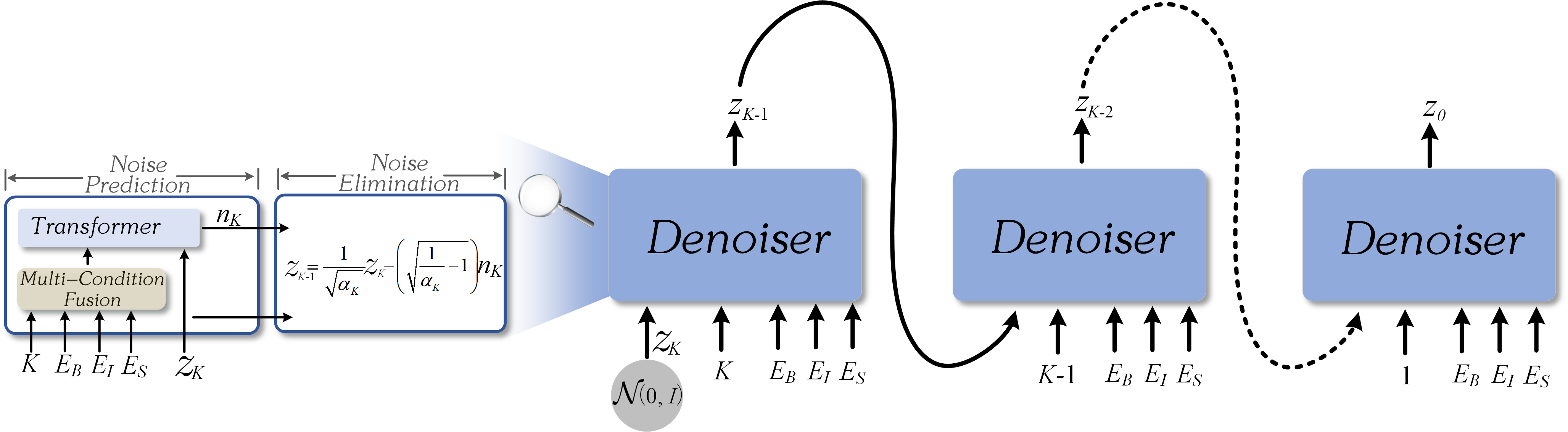}
		\caption{Iterative Denoising Module. Given $\boldsymbol{E}_{B}, \boldsymbol{E}_{I}$ and $\boldsymbol{E}_{S}$ as condition inputs, MCLD learns a conditional denoiser to recursively infer the latent future motion embedding codes $z^{\prime}_{0}$ from a Gaussian noise signal $z_{K}$ with $K$ Markov denoising steps. Then, $\mathcal{D}$ decodes it to 3D motion sequence $\widehat{B}^{+}$.}
		\label{denoising}
	\end{figure*}

		\begin{figure}[t]
		\centering
		\includegraphics[width=0.47\textwidth]{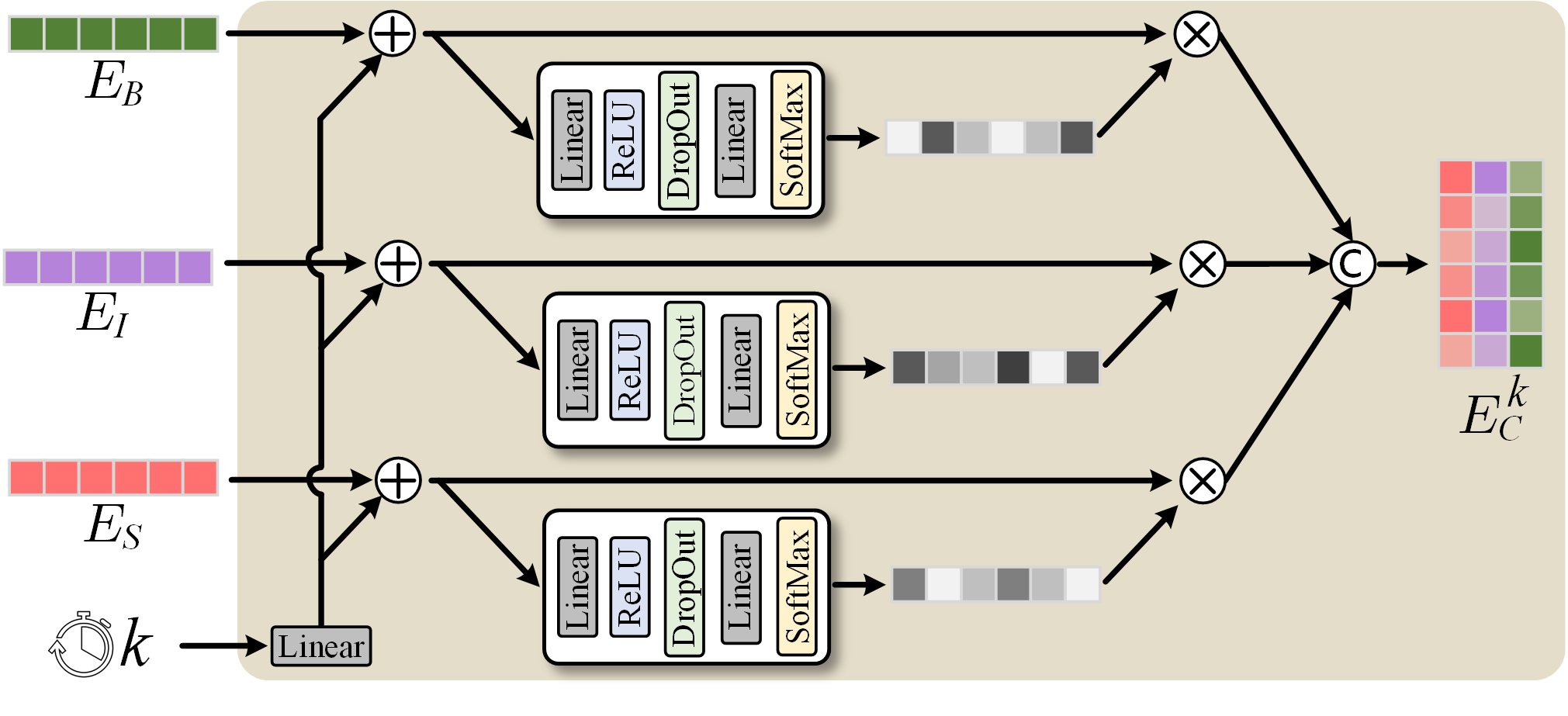}
		\caption{Multi-Condition Fusion Module. MCF dynamically integrates $E_{S}$, $E_{I}$ and $E_{B}$ at $k$-th denoising step in the diffusion model. $\oplus$, $\otimes$, and \textcircled{c} denote addition, element-wise product, and concatenation operations, respectively.}
		\label{mcf}
	\end{figure}

	\subsection{Multi-Attention Encoding}
	\label{method}
	Given the 3D key scene region $S^{\prime}$ and 3D human motion history $B^{-}$ inside it, there are multiple feature representations within these body and scene points, including body motion, scene geometry and body-scene interaction. In this case, we deploy a transformer-based multi-attention encoder (MAE) on $S^{\prime}$ and $B^{-}$ to jointly extract these multiple latent feature embeddings from them. As shown in Fig. \ref{mae_pipeline}, MAE consists of a series of self-attention and cross-attention transformer layers. Specifically, MAE uses a self-attention transformer to extract body motion patterns and scene geometry structures within $B^{-}$ and $S^{\prime}$, respectively. Moreover, it deploys a cross-attention transformer to extract body-scene interaction patterns between $B^{-}$ and $S^{\prime}$. For simplicity, we first take the generic self-attention formulation in scene transformer as an example and introduce the technique defined in each layer. It is similar to the operation in body transformer deployed on $B^{-}$. Given the scene input $S^{\prime}$, we apply linear projection layers to it and transform it into the query $Q_{S}$, key $K_{S}$, and value $V_{S}$ as:
	\begin{equation}
		Q_{S}=S^{\prime} W_{Q},\quad K_{S}=S^{\prime} W_{K},\quad V_{S}=S^{\prime} W_{V},
		\label{eq.2}
	\end{equation}
	where $W_{Q}, W_{K}, W_{V}$ are learnable projection parameter matrices. Then, \textit{Self-Atten}($\cdot$) is used for extracting geometry structure representations as:     
	\begin{equation}
		\begin{aligned}
			Z_{S} &= \textit{Self-Atten}(Q_{S}, K_{S}, V_{S}) \\
			&= \operatorname{softmax} \left(\frac{Q_{S} (K_{S})^{T}}{\sqrt{d_{K}}}\right) V_{S},
		\end{aligned}
		\label{eq.3}	
	\end{equation}
	where $d_{K}$ is the dimension of $K_{S}$. The term $\left(Q_{S} (K_{S})^{T}\right) V_{S}$ can be intuitively interpreted as wide-range feature aggregation based on pair-wise geometric dependencies between $N_{s}^{\prime}$ scene points. Following \cite{DBLP:conf/nips/VaswaniSPUJGKP17}, it is beneficial to linearly project the queries, keys, and values $h$ times with different learned projections. Hence, \textit{Multi-Head} allows the \textit{Self-Atten}($\cdot$) to jointly attend to information from different representation subspaces as:
	\begin{equation}
		\begin{aligned}
			\boldsymbol{E}_{S} = \textit{Multi-Head}(Q_{S},K_{S},V_{S}) = \operatorname{concat}(Z^{1}_{S},\cdots,Z^{h_{e}}_{S}),
		\end{aligned}
		\label{eq.4}
	\end{equation} 
	where $h_{e}$ is the number of the heads we use. We then employ a residual connection and layer normalization techniques to our self-attention layers. We further apply a feed-forward layer, again followed by a residual connection and a layer normalization following \cite{DBLP:conf/nips/VaswaniSPUJGKP17}. The whole process forms one layer of scene transformer. We stack $L_{e}$ such layers to extract the $C_{e}$-dimensional scene geometry embedding $\boldsymbol{E}_{S} \in \mathbb{R}^{C_{e}} $ from $S^{\prime}$. Similar to the scene transformer, body transformer takes $B^{-}$ as its input and extracts body motion embedding $\boldsymbol{E}_{B} \in \mathbb{R}^{C_{e}}$ within $N_{b}\times T$ body points. 
	
	Besides, as shown in Fig. \ref{mae_pipeline}, taking scene points of $S^{\prime}$ as queries, MAE jointly inputs $S^{\prime}$ and $B^{-}$ into a cross-attention transformer for extracting latent body-scene interaction embedding between them. Similarly, we first apply linear layers to $B^{-}$ and $S^{\prime}$ and transform them into the query $Q_{B}$, key $K_{S}$, and value $V_{S}$ as:
	\begin{equation}
	Q_{B}=B^{-} W_{Q},\quad K_{S}=S^{\prime} W_{K},\quad V_{S}=S^{\prime} W_{V}.
	\label{eq.5}
	\end{equation}
	Taking each body point as a query, \textit{Cross-Atten}($\cdot$) infers the interaction between body-scene point pairs as: 
	\begin{equation}
	\begin{aligned}
		Z_{I} &= \textit{Cross-Atten}(Q_{B}, K_{S}, V_{S}) \\
		&= \operatorname{softmax} \left(\frac{Q_{B} (K_{S})^{T}}{\sqrt{d_{K}}}\right) V_{S}.
	\end{aligned}
	\label{eq.6}	
	\end{equation}
	Similarly, integrating \textit{Multi-Head} scheme with \textit{Cross-Atten}($\cdot$) enhances the representation capacity of body-scene interaction embedding $\boldsymbol{E}_{I}$ as:
	\begin{equation}
	\begin{aligned}
	\boldsymbol{E}_{I} = \textit{Multi-Head}(Q_{B},K_{S},V_{S})= \operatorname{concat}(Z^{1}_{I},\cdots,Z^{h_{e}}_{I}). 
	\end{aligned}
	\label{eq.7} 
	\end{equation}
	We also feed them to the normalization and feed-froward layers same as self-attention transformer layers. After applying $L_{e}$ such layers on body points $B^{-}$ and scene points $S^{\prime}$, we extract the latent body-scene interaction embedding $\boldsymbol{E}_{I}\in \mathbb{R}^{C_{e}}$ between them.

    \subsection{Multi-Condition Latent Diffusion}
     Different from other generative frameworks, such as GAN and VAE, the denoising-based diffusion model \cite{DBLP:conf/icml/Sohl-DicksteinW15} learns the cross-modal mapping with the noise prediction strategy. Specifically, a target latent-based future human motion embedding $z$ is gradually noised by the diffusion process. Then, with the joint condition of historical body motion embedding $\boldsymbol{E}_{B}$, current scene embedding $\boldsymbol{E}_{S}$ and body-scene interaction embedding $\boldsymbol{E}_{I}$, a denoiser network learns the reverse diffusion process from denoising the sample step by step. Intuitively, based on the noising-denoising learning strategy, the latent diffusion model learns the probabilistic mapping from the joint conditions of $\boldsymbol{E}_{B}$, $\boldsymbol{E}_{S}$ and $\boldsymbol{E}_{I}$ to $z$. In the following section, we elaborate on the technical details of this conditional latent diffusion model.  

    \textbullet \ \textit{Forward Noising Diffusion.} Specifically, the probabilistic diffusion in the latent space is modeled as a Markov noising process as:
	\begin{equation}
		q\left(z_{k} \mid z_{k-1}\right)=\mathcal{N}\left(\sqrt{\alpha_{k}} z_{k-1},\sqrt{1-\alpha_{k}} I\right),
		\label{eq.8}
	\end{equation}
	where $z_{k}$ denotes the latent motion representation $z$ at $k$-th Markov noising step, $\alpha_{k} \in \left(0, 1\right)$ is a hyper-parameters for the $k$-step sampling. Intuitively, Eq. \ref{eq.8} can be interpreted as sampling a noise from $\epsilon \sim \mathcal{N}(0,1)$ and then injecting it into $z_{k-1}$. Finally, a $K$-length Markov noising process gradually injects random noise into the latent motion representation $z$ and arrives at a noising sequence $\{z_{k}\}_{k=0}^{K}$. If $K$ is sufficiently large, $z_{K}$ will approximate a normal Gaussian noise signal. 
 
    \textbullet \ \textit{Dynamic Multi-Condition Fusion.} \revise{Furthermore, we develop a multi-condition fusion module (MCF) to dynamically integrate $\boldsymbol{E}_{B}$, $\boldsymbol{E}_{S}$ and $\boldsymbol{E}_{I}$ into a joint condition embedding $\boldsymbol{E}_{C}$. Specifically, MCF respectively infers the attention response of $\boldsymbol{E}_{B}$, $\boldsymbol{E}_{S}$ and $\boldsymbol{E}_{I}$ on each input sample and its each denoising diffusion step. For simplicity, we take the dynamic multi-condition fusion operation at $k$-th diffusion step as an example and elaborate on its technical details. As shown in Fig.\ref{mcf}, we first deploy a linear projection $\theta(\cdot)$ to map $k$ into a latent-based embedding and respectively introduce it into $\boldsymbol{E}_{B}$, $\boldsymbol{E}_{S}$ and $\boldsymbol{E}_{I}$ as}:
 	\begin{equation}
  	\begin{aligned}
\widetilde{\boldsymbol{E}}_{B}^k &=\boldsymbol{E}_{B} + \theta(k), \\ \widetilde{\boldsymbol{E}}_{S}^k &=\boldsymbol{E}_{S} + \theta(k), \\
\widetilde{\boldsymbol{E}}_{I}^k &=\boldsymbol{E}_{I} + \theta(k).
		\label{eq.8_1}
  	\end{aligned}
	\end{equation}
\revise{Then, given $\widetilde{\boldsymbol{E}}_{B}^k$, $\widetilde{\boldsymbol{E}}_{S}^k$, and $\widetilde{\boldsymbol{E}}_{I}^k$, MCF dynamically infers their channel-wise attention responses at $k$-th diffusion step as:}
 	\begin{equation}
  	\begin{aligned}
\widehat{\boldsymbol{E}}_{B}^k &=\widetilde{\boldsymbol{E}}_{B}^k \otimes \text{SoftMax}(\theta_B^2(\sigma(\theta_B^1(\widetilde{\boldsymbol{E}}_{B}^k)))), \\ \widehat{\boldsymbol{E}}_{S}^k &=\widetilde{\boldsymbol{E}}_{S}^k \otimes \text{SoftMax}(\theta_S^2(\sigma(\theta_S^1(\widetilde{\boldsymbol{E}}_{S}^k)))),  \\
\widehat{\boldsymbol{E}}_{I}^k &=\widetilde{\boldsymbol{E}}_{I}^k \otimes \text{SoftMax}(\theta_I^2(\sigma(\theta_I^1(\widetilde{\boldsymbol{E}}_{I}^k)))), 
		\label{eq.8_2}
  	\end{aligned}
	\end{equation}
\revise{where $\otimes$ denotes element-wise multiplication, and $\sigma(\cdot)$ is a ReLU-based non-linearity activation function. Finally, at $k$-th multi-condition fusion step, we integrate $\widehat{\boldsymbol{E}}_{B}^k$, $\widehat{\boldsymbol{E}}_{S}^k$, and $\widehat{\boldsymbol{E}}_{I}^k$ into a joint condition embedding $\boldsymbol{E}_{C}^k$ via a channel-wise concatenation as}:
    \begin{equation}
    \boldsymbol{E}_{C}^k = [\widehat{\boldsymbol{E}}_{B}^k, \widehat{\boldsymbol{E}}_{S}^k, \widehat{\boldsymbol{E}}_{I}^k].
 \end{equation}

 \textbullet \ \textit{Reverse Iterative Denoising.} ] \revise{To learn an effective mapping from the conditional embedding $\boldsymbol{E}_{C}$ to the target future human motion embedding $z$, we develop a transformer-based denoiser $\mathcal{R}$ that iteratively anneals the noise of $\{z_{k}\}_{k=0}^{K}$ to infer the reconstructed future human motion embedding $z^{\prime}$.} Specifically, as shown in Fig. \ref{denoising}, the denoising process at $k$-th diffusion step can be factorized into two sequential stages: \textit{Noise Prediction} and \textit{Noise Elimination}. Firstly, at the \textit{Noise Prediction} stage, a Transformer-based denoiser $\mathcal{R}(\cdot)$ takes $\boldsymbol{E}_{C}^k$ and $z_{k}$ as its input and infers the noise signal injected at $k$-th Markov diffusion step and denoises $z_{k}$ as:
 \begin{equation}
		z_{k-1}=\frac{1}{\sqrt{\alpha}_k} z_{k} - \sqrt{\frac{1}{\alpha_{k}}-1} \mathcal{R}(z_{k}, \boldsymbol{E}_{C}^k), 
		\label{eq.9}
	\end{equation}
 Recurring these denoising steps $K$ times, based on the given condition embedding input $\boldsymbol{E}_{C}$, we reconstruct latent future motion embedding $z^{\prime}_{0}$ from the Gaussian noise signal $z_{K}$. Finally, we optimize all these components ($i.e.,$ KRP, MAE, and denoiser) end-to-end, and the training loss is defined as the noise prediction error at the $k$-th Markov denoising step as:
	\begin{equation}
		\mathcal{L}_{\mathcal{R}}  = \mathbb{E} \left[\left\|\epsilon-\mathcal{R}(\boldsymbol{E}_{C}^k,k,z_{k})\right\|_2^2\right] 
		\label{eq.10}
	\end{equation} 
	
	In the inference stage, the trained latent diffusion model infers the future motion embedding $z^{\prime}_{0}$ from the extracted historical body motion embedding $\boldsymbol{E}_{B}$, body-scene interaction embedding $\boldsymbol{E}_{I}$, and current scene geometry embedding $\boldsymbol{E}_{G}$. Then, given $z^{\prime}_{0}$,the decoder $\mathcal{D}$ maps it into the original 3D pose space and generates $\widehat{B}^{+}$ for future human motion prediction.

		\begin{table*}[!t]
		\caption{Comparisons between our method and prior works in terms of pipeline, accuracy, and efficiency. We discuss their pipelines from method input and method architecture and report their model sizes to reflect their computational efficiency. *: For fair comparison, point clouds are reconstructed from corresponding RGB videos. As for non-deterministic diverse prediction, we repeat their evaluations 20 times and report the average with 95\% confidence interval. The best performances are indicated with \textbf{bold}.}
		\centering
		\resizebox{0.99\textwidth}{!}{
			\begin{tabular}{lcccccccc}
				\toprule
				\multicolumn{1}{l|}{\multirow{2}{*}{\textbf{Method}}} & \multicolumn{2}{c|}{\textbf{Method Input}} & \multicolumn{1}{c|}{\multirow{2}{*}{\textbf{Method Architecture}}} & \multicolumn{1}{c|}{\multirow{2}{*}{\# \textbf{Paras}. (M)}} & \multicolumn{2}{c|}{\textbf{GTA-IM}} & \multicolumn{2}{c}{\textbf{PROX}}                       \\
				\multicolumn{1}{c|}{} & Body & \multicolumn{1}{c|}{Scene} & \multicolumn{1}{c|}{}  & \multicolumn{1}{l|}{} & FDE \newrevise{$\downarrow$} & \multicolumn{1}{l|}{ADE \newrevise{$\downarrow$}} & FDE \newrevise{$\downarrow$} & \multicolumn{1}{l}{ADE \newrevise{$\downarrow$}} \\ \midrule
				GPP-Net \cite{DBLP:conf/eccv/CaoGMCVM20} 				& Heatmap &  $\times$    & \textit{CNN}-based VAE                     		& 2.3      & 280 & 192 & 348 & 272   \\ 
				SocialPool \cite{DBLP:journals/ral/AdeliARNR20} 		& Skeleton &  $\times$   & \textit{GRU}-based VAE                        & 1.4      & 218 & 172 & 283 & 254    \\
				Bihmp-GAN\cite{kundu2019bihmp} & Skeleton &  $\times$   & \textit{RNN}-based GAN   & 2.3  & 204 & 171 & 279 & 250    \\
				Skeleton-Graph \cite{DBLP:journals/corr/abs-2109-10257} & Skeleton & $\times$    & \textit{GCN}-based VAE                     		& 1.2    & 194 & 170 & 272 & 247    \\
				\minorrevise{DMMGAN} \cite{nikdel2023dmmgan} & Skeleton & $\times$    & \textit{Transformer}-based GAN & 2.5    & 199 & 174 & 280 & 253    \\
				CA-HMF \cite{DBLP:conf/neurips/abs-2210-03954} & Skeleton &  $\times$ & \textit{GRU}-based VAE  & 5.1  & 186  & 159  &  265  & 239  \\
				\minorrevise{MotionFlow} \cite{zand2023flow} & Skeleton &  $\times$ & \textit{CNN}-based VAE & 4.6 & 182  & 157  & 263  &  231  \\
				\minorrevise{VADER} \cite{yasar2023vader} & Skeleton &  $\times$ & \textit{GRU}-based GAN  & 5.4 & 175  & 152  & 259  & 228   \\
				\minorrevise{EAI} \cite{ding2024expressive} & Skeleton &  $\times$ & \textit{GCN}-based VAE  & 3.8 & 169  & 147  & 253  & 224   \\
				\minorrevise{Actformer} \cite{xu2023actformer} & Skeleton &  $\times$ & \textit{Transformer}-based GAN  & 4.4  & 162  & 148  &  247  & 223 \\
				STAG \cite{scofano2023staged} & Skeleton & $\times$ &  \textit{GRU}-based VAE & 3.4  & 159 & 141 & 251  & 228  \\ 
				\textbf{MCLD} (w/o $\boldsymbol{E}_{S}$, $\boldsymbol{E}_{I}$) 					& Skeleton & $\times$  & \textit{One-Condition} Latent Diffusion  & 4.1 & \textbf{152} & \textbf{129} & \textbf{241} & \textbf{220} \\ \midrule
				GPP-Net \cite{DBLP:conf/eccv/CaoGMCVM20} 				& Heatmap & RGB Image   & \textit{CNN}-based VAE      		& 5.4     & 245 & 181 & 340 & 261  \\ 
				SocialPool \cite{DBLP:journals/ral/AdeliARNR20} 	    & Skeleton & RGB Video  & \textit{GRU}-based VAE      & 3.5    & 203 & 169 & 271  & 243  \\ 
				Skeleton-Graph \cite{DBLP:journals/corr/abs-2109-10257} & Skeleton & RGB Video  & \textit{GCN}-based VAE 		& 3.1    & 190 & 161 & 264  & 240  \\
				SAMP \cite{DBLP:conf/iccv/HassanCVSYZB21}  & Skeleton & RGB Video  & \textit{LSTM}-based VAE 		& 5.5    & 144 & 126 & 245  & 225  \\
				CA-HMF \cite{DBLP:conf/neurips/abs-2210-03954} & Skeleton & Point Cloud$^{*}$\!\!\! &  \textit{GRU}-based VAE & 9.1    & 133 & 117 & 229  & 197  \\ 
				STAG \cite{scofano2023staged} & Skeleton & Point Cloud$^{*}$\!\!\! &  \textit{GRU}-based VAE & 6.5  & 130 & 113 & 216  & 183  \\ 
				\textbf{MCLD}(ours)					& Skeleton  & Point Cloud$^{*}$\!\!\! & \textit{Multi-Condition} Latent Diffusion & 7.5 & \textbf{96} & \textbf{88} & \textbf{192} & \textbf{164}\\
				\bottomrule
			\end{tabular}
		}
		\label{table 1}
	\end{table*}
	
			\begin{table*}[t]
		\caption{Comparisons between our method and prior works on GTA-IM dataset. Their prediction performances are analyzed from predicting 0.5$\sim$2s 3D poses and paths. Similarly, as a non-deterministic prediction system, we repeat the evaluation of our MCLD 20 times and report their average with 95\% confidence interval.}
		\centering
		\resizebox{0.99\textwidth}{!}{
			\begin{tabular}{l|cccccccccc}
				\toprule
				\multirow{3}{*}{\textbf{Method}} & \multicolumn{10}{c|}{\textbf{GTA-IM}}                                                        \\
				& \multicolumn{5}{c|}{3D Pose Error \newrevise{$\downarrow$} (mm)}         & \multicolumn{5}{c|}{3D Path Error \newrevise{$\downarrow$} (mm)} \\
				& \qquad 0.5s \qquad & \qquad 1s \qquad &\qquad 1.5s \qquad &\qquad 2s \qquad & \multicolumn{1}{c|}{\qquad 3s \qquad} & \qquad 0.5s \qquad & \qquad 1s \qquad &\qquad 1.5s \qquad &\qquad 2s \qquad & \multicolumn{1}{c|}{\qquad 3s \qquad} \\ \midrule
				GPP-Net \cite{DBLP:conf/eccv/CaoGMCVM20} 				& \qquad 90 \qquad  & \qquad 131 \qquad & \qquad 161 \qquad & \qquad 193 \qquad & \qquad 291 \qquad &\qquad 96 \qquad & \qquad 135 \qquad & \qquad 216 \qquad & \qquad 264 \qquad & \qquad 425 \qquad \\ 
				SocialPool \cite{DBLP:journals/ral/AdeliARNR20} 	    & \qquad 100 \qquad  & \qquad 112 \qquad & \qquad 136 \qquad & \qquad 142 \qquad & \qquad 276 \qquad & \qquad 103 \qquad & \qquad 140 \qquad & \qquad 191 \qquad & \qquad 248 \qquad & \qquad 411 \qquad \\ 
				Skeleton-Graph \cite{DBLP:journals/corr/abs-2109-10257} & \qquad 87 \qquad & \qquad 101 \qquad & \qquad 119 \qquad & \qquad 133 \qquad & \qquad 234 \qquad & \qquad 91 \qquad & \qquad 142 \qquad & \qquad 176 \qquad & \qquad 241 \qquad & \qquad 397 \qquad  \\
				CA-HMF \cite{DBLP:conf/neurips/abs-2210-03954}          & \qquad 51 \qquad & \qquad 68 \qquad & \qquad 76 \qquad & \qquad 87 \qquad & \qquad  117 \qquad & \qquad 58 \qquad & \qquad 103 \qquad & \qquad 155 \qquad & \qquad 221 \qquad & \qquad 363 \qquad \\
				STAG \cite{scofano2023staged}         & \qquad 48 \qquad & \qquad 66 \qquad & \qquad 73 \qquad & \qquad 82 \qquad & \qquad 103 \qquad & \qquad 55 \qquad & \qquad 100 \qquad & \qquad 147 \qquad & \qquad 208 \qquad & \qquad 349 \qquad \\
				\textbf{MCLD} (ours)            &\qquad \textbf{40} \qquad & \qquad \textbf{57} \qquad & \qquad \textbf{62} \qquad &\qquad \textbf{78} \qquad &\qquad \textbf{89} \qquad &\qquad \textbf{42} \qquad & \qquad \textbf{91} \qquad &\qquad \textbf{128} \qquad &\qquad \textbf{187} \qquad &\qquad \textbf{331} \qquad \\ 
				\bottomrule
			\end{tabular}
		}
		\label{table_2_gta}
	\end{table*}

			\begin{table*}[t]
	\caption{Comparisons between our method and prior works on PROX dataset. Their prediction performances are analyzed from predicting 0.5$\sim$2s 3D poses and paths. We also repeat the evaluation of our MCLD 20 times and report their average with 95\% confidence interval.}
	\centering
	\resizebox{0.99\textwidth}{!}{
		\begin{tabular}{l|cccccccccc}
			\toprule
			\multirow{3}{*}{\textbf{Method}} & \multicolumn{10}{c|}{\textbf{PROX}}                                                       \\
			& \multicolumn{5}{c|}{3D Pose Error \newrevise{$\downarrow$} (mm)}         & \multicolumn{5}{c|}{3D Path Error \newrevise{$\downarrow$} (mm)}\\
			 & \qquad 0.5s \qquad & \qquad 1s \qquad & \qquad 1.5s \qquad & \qquad 2s \qquad & \multicolumn{1}{c|}{\qquad 3s \qquad} & \qquad 0.5s \qquad & \qquad 1s \qquad & \qquad 1.5s \qquad & \qquad 2s \qquad & \multicolumn{1}{c|}{\qquad 3s \qquad}  \\ \midrule
			GPP-Net \cite{DBLP:conf/eccv/CaoGMCVM20} 				& \qquad 141 \qquad & \qquad 241 \qquad & \qquad 369 \qquad & \qquad 491 \qquad & \qquad 778 \qquad & \qquad 131 \qquad & \qquad 203 \qquad & \qquad 263 \qquad & \qquad 342 \qquad & \qquad 507 \qquad \\ 
			SocialPool \cite{DBLP:journals/ral/AdeliARNR20} 	    & \qquad 129 \qquad & \qquad 216 \qquad & \qquad 344 \qquad & \qquad 473 \qquad & \qquad 731 \qquad & \qquad 119 \qquad & \qquad 178 \qquad & \qquad 239 \qquad & \qquad 307 \qquad & \qquad 453 \qquad \\ 
			Skeleton-Graph \cite{DBLP:journals/corr/abs-2109-10257} & \qquad 121 \qquad & \qquad 205 \qquad & \qquad 327 \qquad & \qquad 455 \qquad & \qquad 676 \qquad & \qquad 112 \qquad & \qquad 169 \qquad & \qquad 203 \qquad & \qquad 291 \qquad & \qquad 395 \qquad \\
			CA-HMF \cite{DBLP:conf/neurips/abs-2210-03954}          & \qquad 93 \qquad & \qquad 187 \qquad & \qquad 284 \qquad & \qquad 381 \qquad & \qquad 584 \qquad & \qquad 90 \qquad & \qquad 128 \qquad & \qquad 149 \qquad & \qquad 168 \qquad & \qquad 217 \qquad \\
			STAG \cite{scofano2023staged}    & \qquad 88 \qquad & \qquad 181 \qquad & \qquad 253 \qquad & \qquad 362 \qquad & \qquad 551 \qquad & \qquad 87 \qquad & \qquad 119 \qquad & \qquad 138 \qquad & \qquad 151 \qquad & \qquad 197 \qquad \\
		\textbf{MCLD} (ours) & \qquad \textbf{81} \qquad &\qquad \textbf{173} \qquad &\qquad \textbf{231} \qquad & \qquad \textbf{326} \qquad &\qquad \textbf{521} \qquad  & \qquad \textbf{81} \qquad &\qquad \textbf{101} \qquad & \qquad \textbf{120} \qquad & \qquad \textbf{139} \qquad &\qquad \textbf{179} \qquad \\ 
			\bottomrule
		\end{tabular}
	}
	\label{table_2_prox}
\end{table*}

	\section{Experiments}
	\label{Experiment}
	\subsection{Datasets}
	\label{dataset}
	\textbf{GTA-IM.} GTA Indoor Motion \cite{DBLP:conf/eccv/CaoGMCVM20} is a common-used large-scale dataset that emphasizes human-scene interactions. The synthetic data of motion and interactions were collected from Grand Theft Auto (GTA) gaming engine with control over different actors, scenes, motions. The dataset contains 50 human characters acting inside 10 different large indoor scenes. Each scene has several floors, including living rooms, bedrooms, kitchens, balconies, etc., enabling diverse scene layouts and  human-scene interactions. The collected actions include climbing the stairs, turning a corner, opening the door, lying down, sitting, etc. -- a set of daily indoor activities. Besides, these actors have 22 walk styles. All of these factors enable the motion data to be more realistic and the indoor 3D human motion prediction task more challenging. In total, GTA-IM collects 1M RGBD frames with ground truth 3D human pose (21 joints) to form diverse action videos at 5 FPS. Following the common-used setups \cite{DBLP:journals/corr/abs-2109-10257,DBLP:conf/eccv/CaoGMCVM20}, we split 8 scenes for training and rest 2 scenes for testing.
	
	\textbf{PROX.} Proximal Relationships with Object eXclusion \cite{DBLP:conf/iccv/HassanCTB19} is a real-world human motion dataset captured using Kinect-One sensor. It employs sensors to scan 12 indoor scenes and capture 20 subjects moving in and interacting with these scenes. These scenes can be grouped to: bedrooms, living rooms, offices, etc. In total, PROX collects 0.1M RGBD frames with ground truth 3D human pose (18 joints). Notably, due to the extensive efforts required for manual collecting and real-world acting, PROX has a relatively small number of activities and characters. Besides, real sensors limit the range of indoor scenes and human movements. In these cases, models trained on this dataset tend to be overfitting to the training data. Hence, the performance on the PROX shows the robustness of our method. Following the setups of \cite{DBLP:journals/corr/abs-2109-10257,DBLP:conf/eccv/CaoGMCVM20}, we down-sample PROX videos to 5 FPS and split them with 52 training sequences and 8 sequences for testing. \textit{Please refer to the appendix for more details on datasets and data preparation.}
	
	\begin{figure*}[t]
		\centering
		\includegraphics[width=1\textwidth]{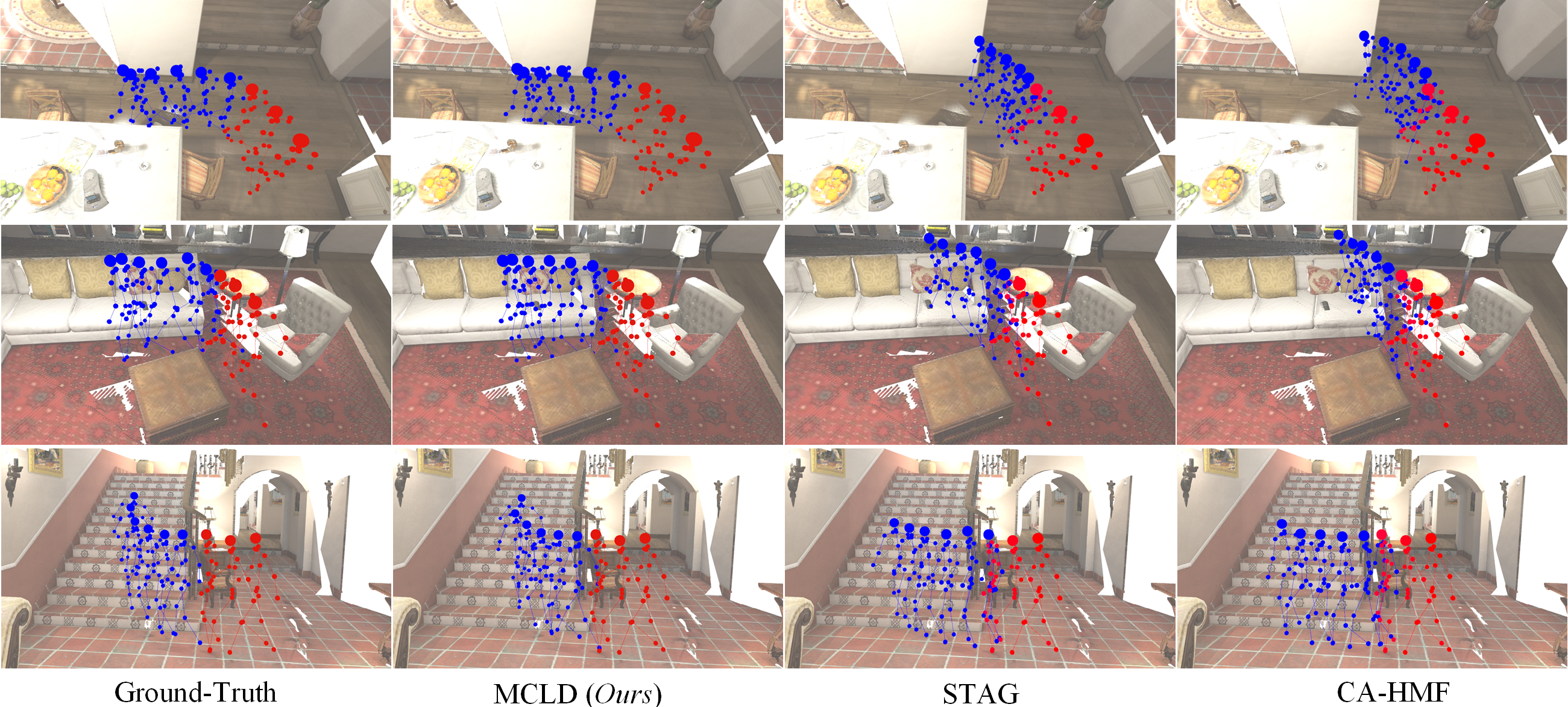}
		\caption{Qualitative Comparison. Given the 3D human motion history (red skeletons) and 3D scene point cloud as inputs, we visualize the future human motion prediction results (blue skeletons) of STAG \cite{scofano2023staged}, CA-HMF \cite{DBLP:conf/neurips/abs-2210-03954}  and Our MCLD. Compared to other methods, MCLD generates 3D indoor human motions that are more realistic and more compatible with the context of the scene around them.}
		\label{vis_compair}
	\end{figure*}
 
	\subsection{Model Configuration}
	\label{config}
	In our experiments, we give 1-second history ($i.e.$, $T$=5) as input and predict the human motions in future 2 seconds ($i.e.$, $\Delta T$=10). The number of body joints within each skeleton is 21 for GTA-IM and 18 for PROX ($i.e.$, $N_{b}$=18/21). The number of scene points in the key region $S^{\prime}$ is uniformly sampled down to 6000 ($i.e.$, $N_{s}^{\prime} = 6000$). In the future motion embedding, encoder $\mathcal{E}$ and decoder $\mathcal{D}$ are 6-layer transformer with 4 heads ($i.e.,$, $L_{v}=6$). The feature channel of future motion embedding $z$ is 512 ($i.e.,$ $C_{e}=512$). In the multi-condition input embedding, key region proposal module uses a 3-layer transformer with 4 heads ($i.e.$, $L_{k}=3, h_{k}=4$) to extract motion features and deploys a MLP layer to regress range-related parameter set ($i.e., \{O,L,W,H\}$). The multi-attention encoder contain 6 layers with 4 heads ($i.e.$, $L_{e}=6, h_{e}=4$). In the latent diffusion model, the denoiser $\mathcal{R}$ is a 9-layer transformer, and the step number of the Markov noising process is 1000 ($i.e., K=1000$). Finally, we implement MCLD with Pytorch 1.3 on two RTX-3090 GPUs. The parameters of MCLD are optimized with a two-stage training scheme. As for the training of VAE $\mathcal{V}$, its training weight of $\mathcal{L}_{mr}$ and $\mathcal{L}_{kl}$ are set as $1$ and $10^{-4}$, respectively ($i.e.$, $\lambda_{mr}=1, \lambda_{kl}=10^{-4}$).The epochs of VAE and multi-condition latent diffusion training stages are 1k and 4k, respectively. AdamW optimizes all parameters with initial learning rate $10^{-4}$.

	\subsection{Evaluation Metrics}
	\label{metrics}
	Following common methods, we adopt Mean Per Joint Position Error (MPJPE) as a metric for quantitatively evaluating both 3D path and 3D pose prediction. Specifically, we measure the pose error in future $\Delta T$ frames by calculating mean $\ell_{1}$ distant between the $\widehat{B}^{+}_{1: \Delta T}$ and $B^{+}_{1: \Delta T}$. In the case of 3D path error, a specific joint $j$ (such as the center of the skeleton torso) is chosen to reflect the trajectory difference between $\widehat{B}^{+}_{1: \Delta T, j}$ and $B^{+}_{1: \Delta T, j}$. Besides, other two quantitative metrics are also common-used for human motion prediction evaluation \cite{DBLP:conf/cvpr/AlahiGRRLS16,DBLP:conf/cvpr/GuptaJFSA18,DBLP:conf/cvpr/MohamedQEC20}. Mean Average Displacement Error (ADE) reflects the performance errors over all $\Delta T$ frames. Mean Final Displacement Error (FDE) reflects the performance errors within the final ${\Delta T}$-th frame. To comprehensively analysis MCLD, we report its short-term and long-term prediction performances (0.5s $\sim$ 3s) in millimeters (mm) at two large-scale datasets (GTA-IM and PROX). Notably, as a non-deterministic prediction system, we repeat the evaluation of MCLD 20 times and report the average result with 95\% confidence interval for fair comparisons. 

	\subsection{Methods for Comparisons}
	\label{baselines}
	 In this section, we try our best to collect related literature and sketch their main frameworks comprehensively. \minorrevise{We compare our MCLD with recent scene-aware human motion prediction methods, including GPP-Net \cite{DBLP:conf/eccv/CaoGMCVM20}, SocialPool \cite{DBLP:journals/ral/AdeliARNR20}, CA-HMF \cite{DBLP:conf/neurips/abs-2210-03954}, and STAG \cite{scofano2023staged}. Furthermore, we also tune the conditional context input of MCLD from scene-motion conditions to motion-only conditions to compare MCLD with various pose-driven human motion prediction systems, including VAE-based methods \cite{zand2023flow, ding2024expressive, scofano2023staged, DBLP:conf/neurips/abs-2210-03954} and GAN-based methods \cite{kundu2019bihmp, yasar2023vader, nikdel2023dmmgan, xu2023actformer}. Different from these generative methods, MCLD proposes a novel latent-based multi-condition diffusion scheme that learns the future human motion probabilistic distribution jointly conditioned on past body motions and current scene contexts.} Besides, we develop a series of powerful components, including KRP and MAE, significantly improving human motion prediction. In the following experiment section, we analyze the individual components and their configurations in the final architecture.

		\begin{figure*}[t]
		\centering
		\includegraphics[width=1\textwidth]{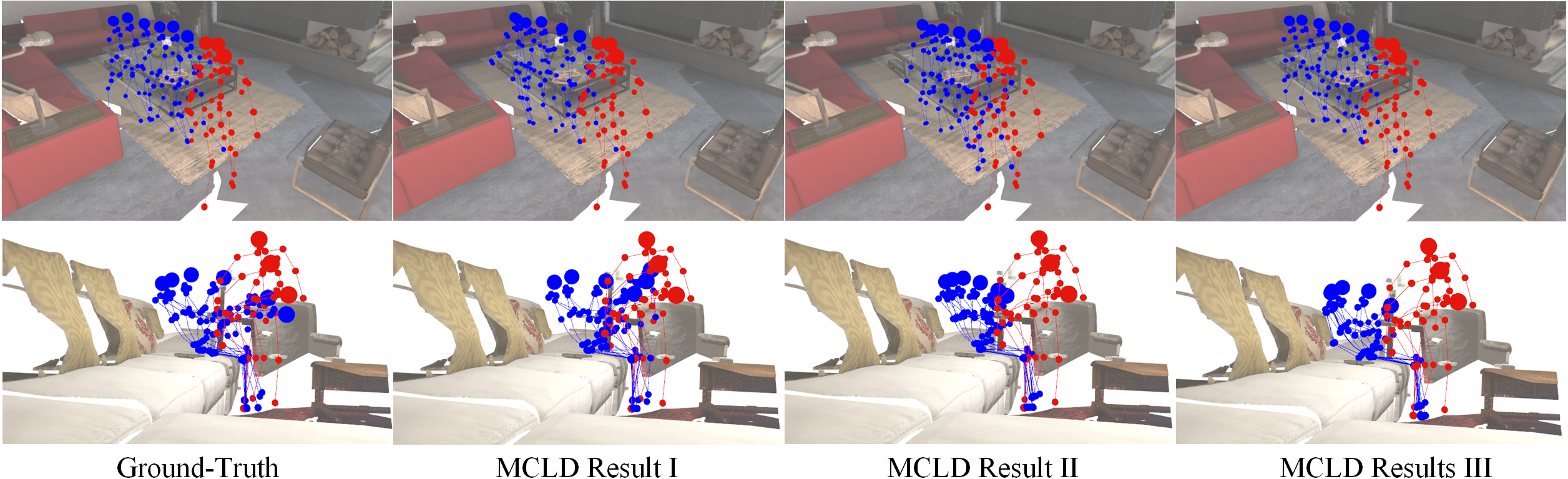}
		\caption{Diverse Predictions. As a non-deterministic prediction system, MCLD is able to generate diverse and reasonable future human motions (blue skeletons) from the same scene point cloud and body motion history input (red skeletons).}
		\label{vis_diversity}
	\end{figure*}
	\textbf{\textit{Quantitative Comparisons.}}
	We provide detailed comparisons between our method and prior works. In Tab.\ref{table 1}, we first compare these models from three aspects: \emph{pipeline}, \emph{efficiency}, and \emph{accuracy}. Then, each aspect is further analyzed from multiple viewpoints. Specifically, the \emph{pipeline} of each model is considered from its input and architecture. To explore the effect of scene input on the final motion prediction in each method, we respectively report their performances with body-only and body-scene input. Their \emph{efficiency} performances are reflected from their number of parameters. As described in Sec.\ref{metrics}, ADE and FDE serve as two \emph{accuracy} metrics to evaluate the quality of prediction. In Tab.\ref{table_2_gta} and Tab.\ref{table_2_prox}, we further compare their 3D pose and path errors on predicting 0.5$\sim$2 seconds motion, respectively. Analyzing the results shown in Tab.\ref{table 1}, Tab.\ref{table_2_gta} and Tab.\ref{table_2_prox}, we have several observations and summarize them into the following: \\
	\textbf{(I)} Diffusion model is a powerful generator for human motion prediction. As shown in Tab. \ref{table 1}, given the body-only input, MCLD clearly improves both short-term and long-term prediction performances over other VAE-based methods on GTA-IM and PROX datasets. For example, MCLD outperforms CA-HMF \cite{DBLP:conf/neurips/abs-2210-03954} on GTA-IM by large margins: 15\% on FDE and 14\% on ADE. These superior prediction performances verify the effectiveness of the diffusion-based generation model on bridging historical and future human motion. \\
	\textbf{(II)} Multi-condition diffusion model is effective in extracting joint priors from multiple condition contexts. As shown in Tab. \ref{table 1}, compared with other works, our model achieves more significant improvement when the scene input is introduced into human motion prediction. We conjecture that there may be two reasons why the scene input brings limited gains to other baseline models: \textbf{a)} The various appearances of the indoor scenes ($e.g.$, different colors, textures and shapes of a same object) make these RGB-based visual patterns ineffective in capturing robust human-scene interaction representations. \textbf{b)} Without a multi-attention feature extractor for inferring body-scene interaction, these baseline methods decouple the body motion modeling and scene context modeling into two separate information flows straightforwardly. \\
	\textbf{(III)} MCLD significantly enhances the stability of long-term prediction. As shown in Tab. \ref{table_2_gta} and \ref{table_2_prox}, we can see that prior works diverge drastically when predicting long-term human poses and trajectories. These results indicate that compared to a short-term prediction, scene context and body-scene interaction are of more vital importance for predicting long-term human motion. In other words, without strong environmental constraints from the current 3D scene context, these methods tend to accumulate prediction errors from one frame to the next and make long-term predictions more divergent as time goes on.

	\textbf{\textit{Qualitative Comparisons.}}
	In this section, we compare the performance of different methods via visualization analyses. As shown in Fig.\ref{vis_compair}, we can see that compare with CA-HMF \cite{DBLP:conf/neurips/abs-2210-03954} and our MCLD, STAG \cite{scofano2023staged} suffers from insufficient contact and collision constraints from steps and chairs, respectively. Furthermore, in contrast to CA-HMF \cite{DBLP:conf/neurips/abs-2210-03954}, the yellow ellipses in Fig.\ref{vis_compair} indicate that MCLD generates realistic human motions that are compatible with its surrounding scene context, significantly improving human motion prediction. For example, MCLD tends to generate freezing body motions at the corner of the staircase. All these quantitative and qualitative comparisons verify that MCLD is a powerful human motion prediction system that outperforms previous methods in terms of realistic and reasonable prediction.
	
	\textbf{\textit{Diverse Prediction.}} As a non-deterministic human motion prediction system, we visualize diverse human motion prediction results inferred from the same human motion history and contextual scene point cloud inputs. As shown in Fig.\ref{vis_diversity}, we respectively visualize three human motion prediction samples of two daily indoor activities. Analyzing these diverse prediction results, we find that their body pose details at each time step have subtle differences while their location movements and motion trajectories tend to be consistent. All these quantitative (Tab.\ref{table_2_gta}) and qualitative performance analyses (Fig.\ref{vis_diversity}) verify that MCLD is a powerful human motion prediction system that significantly improves realistic and diverse motion prediction.  

	\begin{table}[t]
	\caption{Performance comparison between different VAE configurations. The best performances are indicated with bold.}
		\centering
		\resizebox{0.48\textwidth}{!}{
		\begin{tabular}{ccc|cccc}
			\toprule
			\multicolumn{3}{c|}{VAE Configuration}                                                       & \multicolumn{4}{c}{3D Pose Error $\downarrow$ (mm)}                 \\ \hline
			\multicolumn{1}{c|}{$\mathcal{E}$ Layer} & \multicolumn{1}{c|}{$\mathcal{D}$ Layer} & \multicolumn{1}{c|}{$z$ Dimension} & \multicolumn{1}{c|}{\; 0.5s \;} & \multicolumn{1}{c|}{1.0s} & \multicolumn{1}{c|}{\; 1.5s \;} & \multicolumn{1}{c}{\; 2.0s \;} \\ \hline
			 3       &     3                         &          128                      &   49                     &   69                     &  73                      & 91    \\
			 3       &     3                         &          256                      &   47                     &   67                     &  71                      & 89    \\
			 3       &     3                         &          512                      &   47                     &   66                     &  69                      & 87    \\ \hline
			 6       &     6                         &          128                      &   45                     &   64                     &  67                      & 84    \\
			 6       &     6                         &          256                      &   45                     &   62                     &  67                      & 84    \\
			 6       &     6                         &          512                      &   42                     &   60            &  63             & \textbf{78}    \\ \hline
			 9       &     9                         &          128                      &   42                     &   61                     & \textbf{62}              & 82    \\
			 9       &     9                         &          256                      &   \textbf{40}            &   60                     & 66                       & 82    \\
			 9       &     9                         &          512                      &   \textbf{40}            &   \textbf{57}            & \textbf{62}                       & \textbf{78}    \\ \bottomrule
		\end{tabular}}
	\label{abl_vae}
	\end{table}
 
    \subsection{Component Studies}
	\label{abl}
	In this section, we analyze the individual components and their configurations in the final MCLD architecture. Unless stated, the reported performances are ADE results on GTA-IM dataset.
	
	\textbf{\textit{Effects of Latent Motion Representation.}} As presented in Sec.\ref{sec_lmr}, in the first training stage, we deploy a transformer-based VAE $\mathcal{V} = \{ \mathcal{E}, \mathcal{D} \}$ on future human motion to embed it into a low-dimensional feature space and extract its latent motion embedding $z$. In the following, we conduct extensive experiments to verify the effectiveness of this module and investigate its optimal configuration, including the layer of encoder $\mathcal{E}$ and decoder $\mathcal{D}$ and the dimension of latent embedding $z$. Analyzing the results reported in Tab.\ref{abl_vae}, we can see that a 6-layer VAE with 512-dimensional latent embedding brings MCLD the best prediction performance. We conjecture that an over-large encoder $\mathcal{E}$ and decoder $\mathcal{D}$ would make its optimization more difficult. Besides, fewer feature dimensions of the latent embedding code $z$ would make it less representative.

	\begin{figure}[t]
		\centering
		\includegraphics[width=0.48\textwidth]{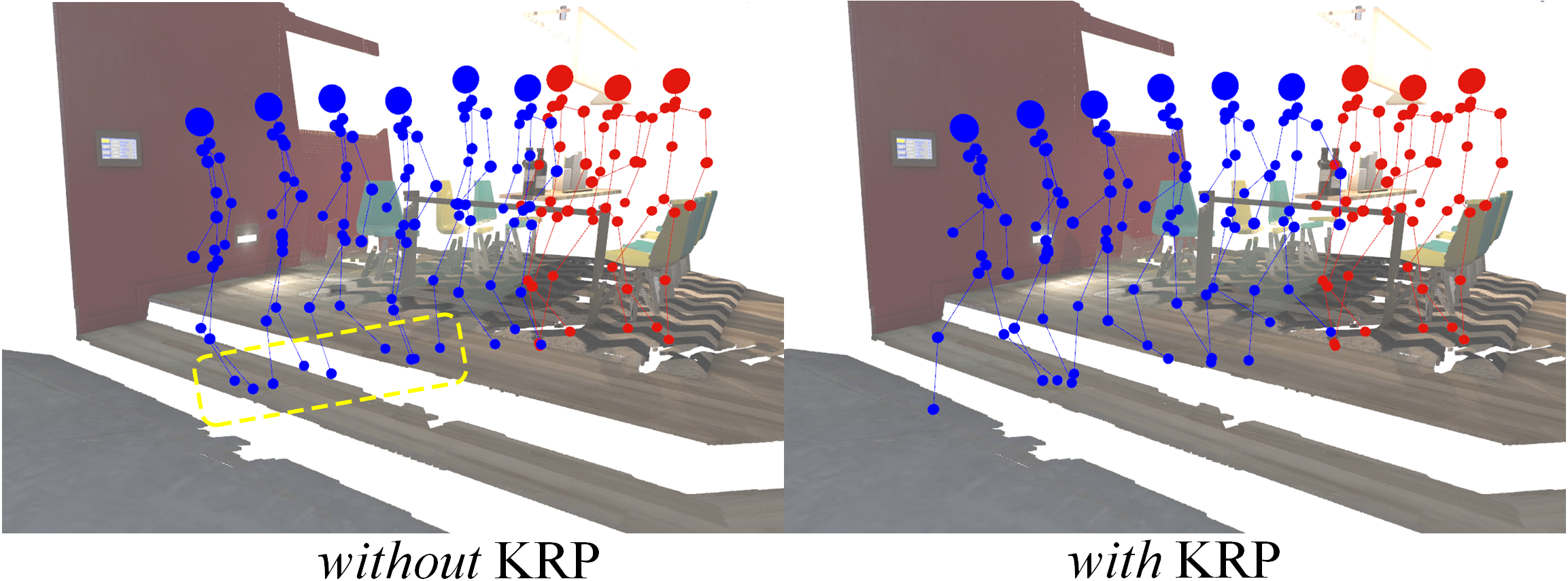}
		\caption{\newrevise{Visualization comparisons between the human motion prediction with and without a KPR module. KRP improves MCLDs with more realistic human-scene interactions (shown in yellow box).}}
		\label{vis_w_wo_krp}
	\end{figure}
	
	\begin{table}[t]
		\caption{Performance comparison between different KRP configurations. The best performances are indicated with bold.}
		\centering
		\resizebox{0.48\textwidth}{!}{
\begin{tabular}{ccc|c|cccc}
	\toprule
	\multicolumn{3}{c|}{\multirow{2}{*}{Key Region Proposal Configuration}}       &  \multirow{2}{*}{\# Paras. (M)}                                                                                       & \multicolumn{4}{c}{3D Pose Error $\downarrow$ (mm)}                \\ \cline{5-8} 
	\multicolumn{3}{c|}{}             &                                                                & \multicolumn{1}{c|}{0.5s} & \multicolumn{1}{c|}{1.0s} & \multicolumn{1}{c|}{1.5s} & 2.0s \\ \hline
	\multicolumn{3}{c|}{w/o KRP}                 &         0                   &             59              &             78              &      89                     &    101  \\ \hline
	\multicolumn{1}{c|}{\multirow{9}{*}{w. KRP}} & \multicolumn{1}{c|}{\multirow{3}{*}{$L_{k}$=3}} & $h_{k}$=4  & 2.1 & 40          & 57               & 62               & 78     \\
	\multicolumn{1}{c|}{}                        & \multicolumn{1}{c|}{}                      & $h_{k}$=6       & 3.8 & 40          & \textbf{56}               & 62               & 78\\
	\multicolumn{1}{c|}{}                        & \multicolumn{1}{c|}{}                      & $h_{k}$=8       & 5.2 & 40          & 58               & \textbf{61}               & \textbf{77}   \\ \cline{2-3}
	\multicolumn{1}{c|}{}                        & \multicolumn{1}{c|}{\multirow{3}{*}{$L_{k}$=5}} & $h_{k}$=4  & 4.0  & \textbf{39}   & 57               & 62               & \textbf{77}     \\
	\multicolumn{1}{c|}{}                        & \multicolumn{1}{c|}{}                      & $h_{k}$=6       & 5.5  &  \textbf{39}         &  57             & 62                & 78     \\
	\multicolumn{1}{c|}{}                        & \multicolumn{1}{c|}{}                      & $h_{k}$=8       & 6.7   &  \textbf{39}         &  59             & 62                & \textbf{77}      \\ \cline{2-3}
	\multicolumn{1}{c|}{}                        & \multicolumn{1}{c|}{\multirow{3}{*}{$L_{k}$=7}} & $h_{k}$=4  & 5.1 & \textbf{39}         &  58             & \textbf{61}                & \textbf{77}      \\
	\multicolumn{1}{c|}{}                        & \multicolumn{1}{c|}{}                      & $h_{k}$=6       & 6.9    & \textbf{39}         &  57             & \textbf{61}                & \textbf{77}    \\
	\multicolumn{1}{c|}{}                        & \multicolumn{1}{c|}{}                      & $h_{k}$=8       & 8.3    & \textbf{39}         &  57             & 62                & \textbf{77}      \\ \bottomrule
\end{tabular}
	}
\label{abl_krp}
	\end{table}

\begin{figure}[t]
	\centering
	\includegraphics[width=0.48\textwidth]{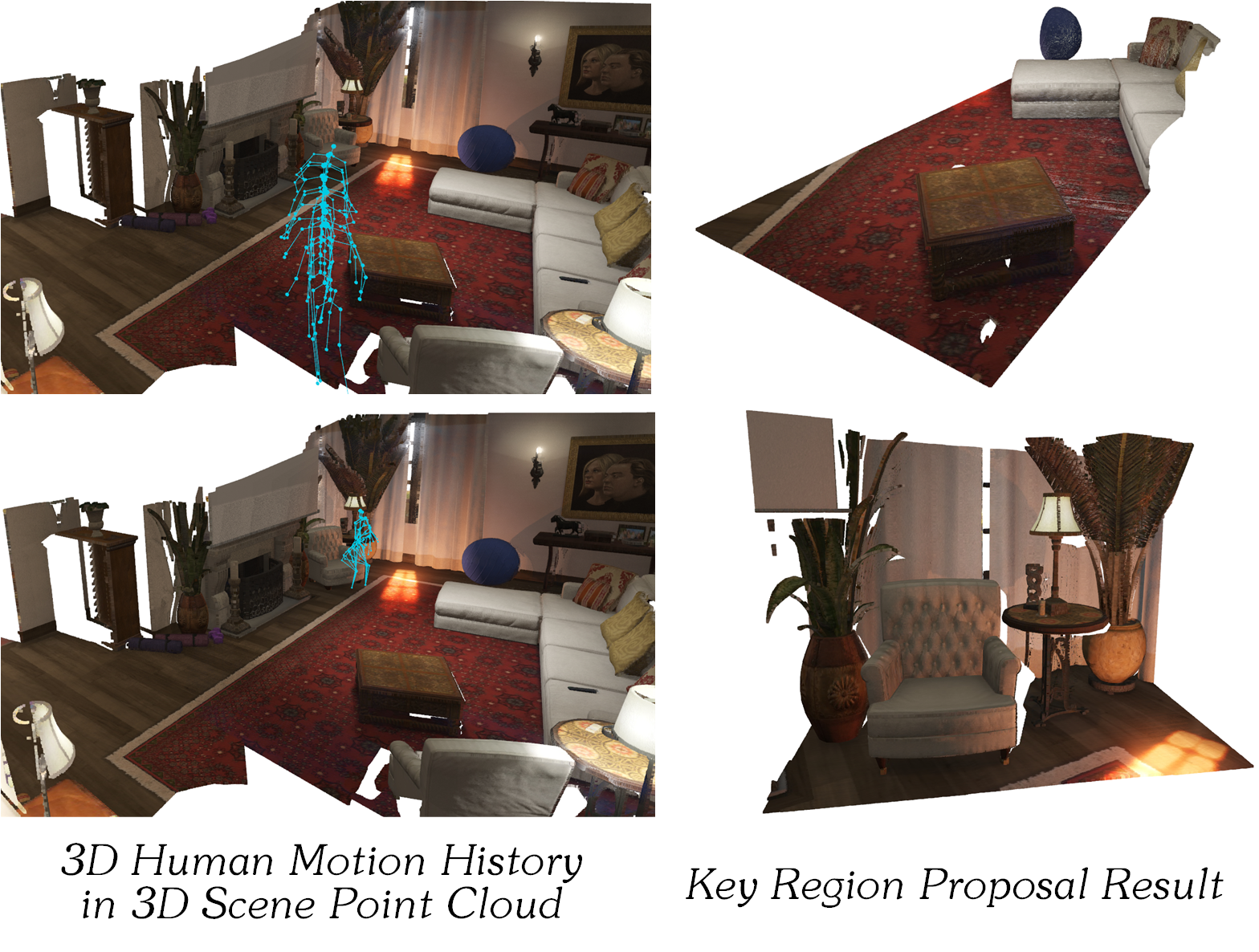}
	\caption{Visualization of key region proposal results. Given two different human motion histories inside the same 3D scene, the KRP module infers two different localized interaction-related regions for these human motion samples.}
	\label{krp_result}
\end{figure}

	\textbf{\textit{Effects of Key Region Proposal.}} \newrevise{In this section, we verify the effectiveness and explore the optimal configurations of the proposed key region proposal (KRP) module with quantitative and qualitative analyses.}
	 \newrevise{Fig.\ref{vis_w_wo_krp} verifies that KRP benefits scene-aware human motion prediction via focusing on an interaction-related localize scene region details, thus significantly improves realistic human-scene interactions.} Furthermore, as shown in Tab.\ref{abl_krp}, we first report the quantitative prediction performance of MCLD without KRP (top row). We can see that integrating KRP into MCLD can significantly improve its quantitative motion prediction performance, proving the effectiveness of this module. It verifies that KRP is beneficial to reduce the redundancy of the initial scene input and isolate MCLD from unnecessary scene noises. Furthermore, we tune the $L_{k}$ and $h_{k}$ configuration of KRP and explore its optimal choices. Analyzing the results shown in Tab.\ref{abl_krp}, we see that KPR is insensitive to the different configuration choices of $L_{k}$ and $h_{k}$, indicating that the generic key region proposal strategy is the main reason for the observed improvements. Therefore, we choose $L_{k}=3$ and $h_{k}=4$ as the default configurations to balance the human motion prediction performance and the computational costs. Besides, we further verify the effectiveness of KRP by analyzing its visualization results. As shown in Fig.\ref{krp_result}, we find that given two different human motion histories inside the same 3D scene, KRP module adaptively infers two different localized interaction-related regions based on human motion patterns. 

\begin{table}[t]
		\caption{Performance comparison between different latent diffusion configurations. The best performances are indicated with bold.}
		\centering
		\resizebox{0.48\textwidth}{!}{
			\begin{tabular}{c|ccc|cccc}
				\toprule
				\multicolumn{4}{c|}{Conditional Latent Diffusion Configuration}                                    & \multicolumn{4}{c}{\multirow{2}{*}{3D Pose Error $\downarrow$ (mm)}} \\ \cline{1-4}
				\multicolumn{1}{c|}{\multirow{2}{*}{$\mathcal{R}$ Layer}} & \multicolumn{3}{c|}{Condition Inputs} & \multicolumn{4}{c}{}                        \\ \cline{2-8} 
				\multicolumn{1}{c|}{}                                     & $\quad \boldsymbol{E}_{B} \quad$   & $\quad \boldsymbol{E}_{I} \quad$   & $\quad \boldsymbol{E}_{S} \quad$  & 0.5s & 1s        & 1.5s          & 2s     \\ \hline
				3   &  $\checkmark$    &  $\checkmark$    &   $\checkmark$                &  52           & 70    &     78               & 98            \\
				6   &  $\checkmark$    &  $\checkmark$    &   $\checkmark$                &  49           & 64     &    71               & 89           \\
				9   &  $\checkmark$    &  $\checkmark$    &   $\checkmark$                &  \textbf{42}  & \textbf{60}   & \textbf{65}  & \textbf{81}        \\ \hline
				\multirow{6}{*}{9}   &  $\checkmark$    &  --    &   --                &  59                  &   79       &  88                  & 105 \\
				   &  --    &  $\checkmark$    &   --                &  63                  &   84       &  95                  & 113  \\
				   &  --    &  --    &   $\checkmark$                &  72                  &   89       &  97                  & 128   \\
			       &  $\checkmark$    &  $\checkmark$   &   --       &  51                  &   73       &   82                 & 94     \\
			       &  $\checkmark$    &  --   &   $\checkmark$       &  49                  &   68       &   77                 & 89  \\
				   &  --    &  $\checkmark$   &   $\checkmark$       &  68                  &   84       &   92                 & 117  \\  \bottomrule
				
			\end{tabular}
		}
		\label{abl_lcd}
	\end{table}
 
\begin{figure}[t]
		\centering
		\includegraphics[width=0.4\textwidth]{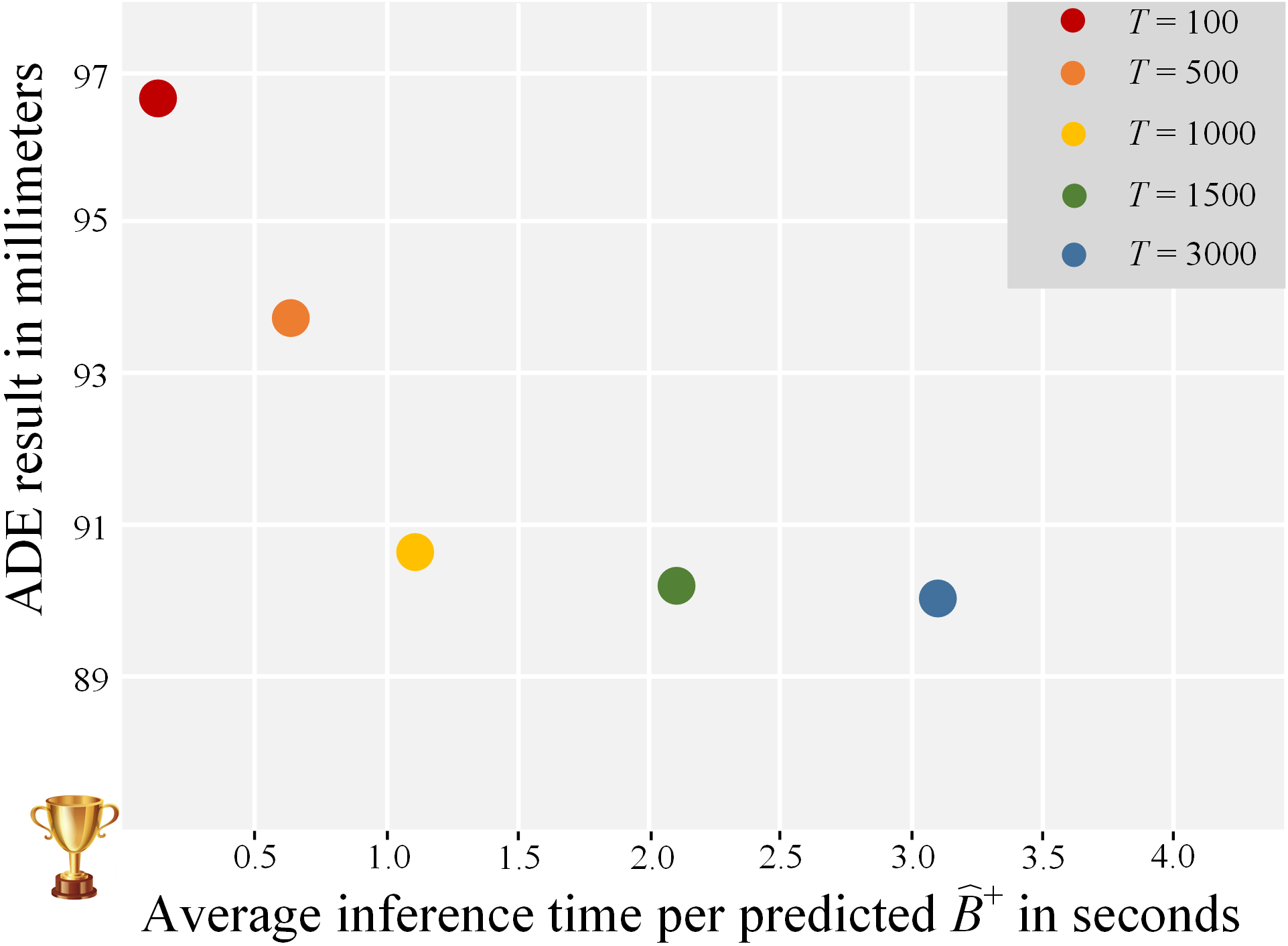}
		\caption{The human motion prediction performance and its average inference time at each configuration choice of Markov denoising step $T$.}
		\label{T_results}
	\end{figure}

	\textbf{\textit{Effects of Latent Conditional Diffusion.}} In the following, we tune the configuration of the proposed conditional latent diffusion model, including the layer of denoiser $\mathcal{R}$, conditional inputs and Markov denoising steps $T$. As shown in Tab.\ref{abl_lcd}, we firstly tune the layer of denoiser $\mathcal{R}$ from 3 to 9. We see that a 9-layer denoiser brings the best prediction performance, indicating that a bigger denoiser significantly improves the denoising-based latent diffusion process. Furthermore, we tune the configuration of condition inputs ($i.e.,$ $\boldsymbol{E}_{B}, \boldsymbol{E}_{I},$ and $\boldsymbol{E}_{S}$) and investigate their effects on future human motion prediction. Specifically, we enumerate 7 different condition compositions of $\boldsymbol{E}_{B}, \boldsymbol{E}_{I},$ and $\boldsymbol{E}_{S}$ and analyze the human motion prediction performance conditioned on each of them. Tab.\ref{abl_lcd} verifies the effectiveness of introducing the joint conditions of $\boldsymbol{E}_{B}, \boldsymbol{E}_{I},$ and $\boldsymbol{E}_{S}$ into the future human motion prediction. Besides, we further explore the effects of the Markov denoising steps $T$ on the denoising-based diffusion process and provide 5 configuration choices for $T$ ($i.e., T \in \{100,500,1000,1500,3000\}$). As shown in \newrevise{Fig.}\ref{T_results}, we also report the average inference time of each prediction at every $T$ configuration to analyze its effects on the computational cost. As shown in Fig.\ref{T_results}, when choosing $T=1500/3000$, the performance gains brought by an over-large $T$ are limited. In this case, we set $T=1000$ to balance the human motion prediction performance and its computational cost.        

\begin{table}[t]
\caption{Performance comparison between different configurations of the MCF module. The best performances are indicated with bold.}
		\centering
		\resizebox{0.48\textwidth}{!}{
\begin{tabular}{cc|cccc}
\toprule
\multicolumn{2}{c|}{\multirow{2}{*}{Multi-Condition Fusion Configuration}} & \multicolumn{4}{c}{3D Pose Error $\downarrow$}                                                                         \\ \cline{3-6} 
\multicolumn{2}{c|}{}                                                      & \multicolumn{1}{c}{0.5s} & \multicolumn{1}{c}{1.0s} & \multicolumn{1}{c}{1.5s} & \multicolumn{1}{c}{2.0s} \\ \hline
\multicolumn{1}{c|}{\multirow{2}{*}{w/o MCF}} & Element-wise Addition      &   48                       &     66                     &     69                    &         87                 \\
\multicolumn{1}{c|}{}                         & Channel-wise Concatenation &    45                      &      63                    &      68                    &        85                  \\ \hline
\multicolumn{1}{c|}{\multirow{2}{*}{w. MCF}}  & w/o embedding of $t$         &  43                        &    61                      &       66                   &        81                  \\
\multicolumn{1}{c|}{}                         & w. embedding of $t$          &  \textbf{40}       & \textbf{57}                       & \textbf{62}                    &    \textbf{78}                   \\
\bottomrule
\end{tabular}
}
\label{MCF}
\end{table}

\begin{figure}[t]
		\centering
		\includegraphics[width=0.4\textwidth]{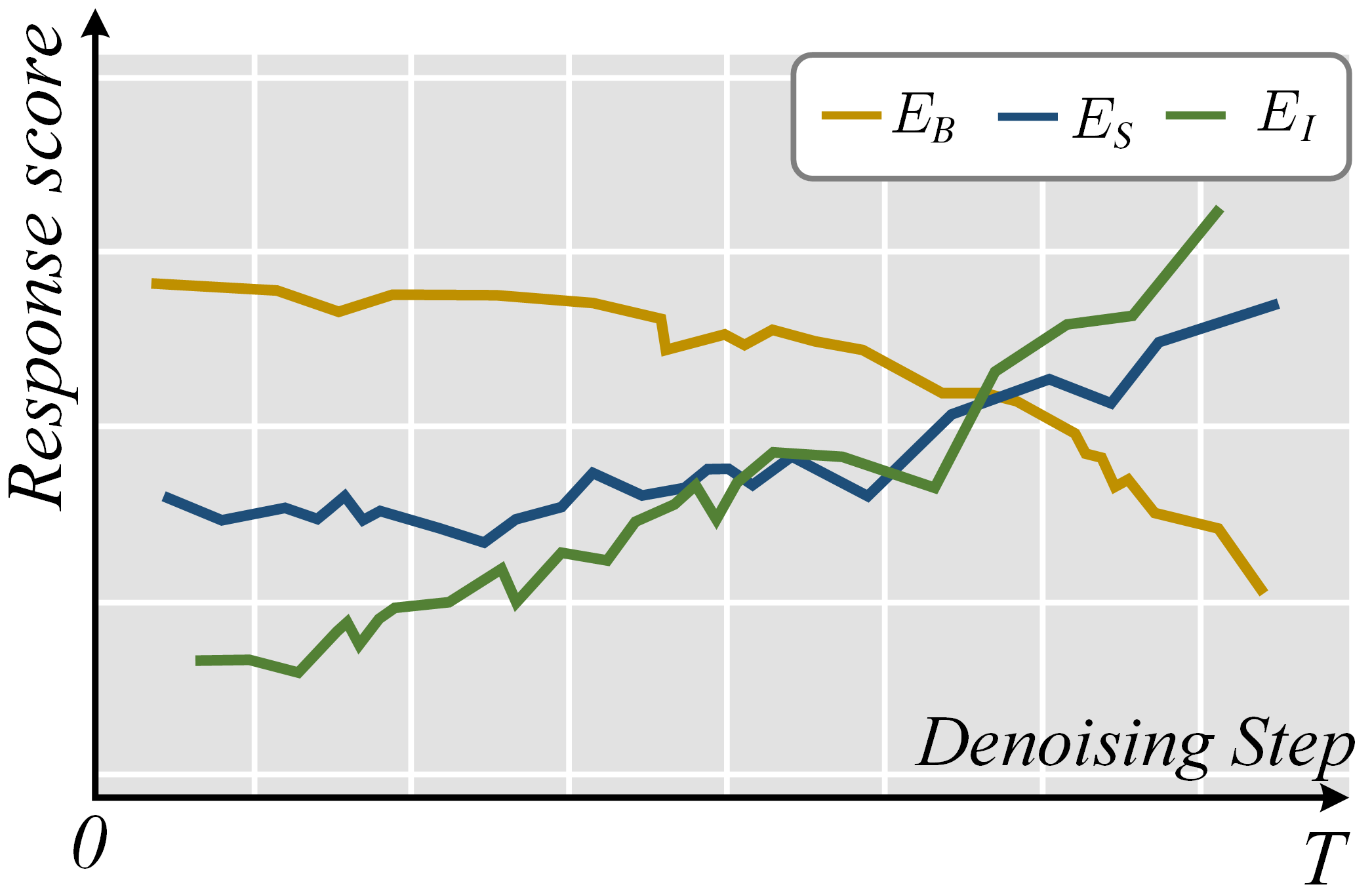}
		\caption{Response scores for different conditions inferred at different diffusion steps.}
		\label{response_scores}
	\end{figure}

    \textbf{\textit{Effects of Dynamic Multi-Condition Fusion.}} \revise{In this section, we analyze the effectiveness of the proposed multi-condition fusion module from two experiments. Firstly, we verify the effect of the MCF module on the human motion prediction performance. As shown in Tab.\ref{MCF}, we can see that (1) introducing MCF into MCLD significantly improves realistic human motion prediction; (2) introducing the embedding of diffusion step $t$ into MCLD also benefits the final performance. Besides, we further investigate the inferred response scores of $E_{B}$, $E_{S}$, and $E_{I}$ to verify the adaptiveness of MCLD over different diffusion steps. Specifically, we average the channel-wise attention scores of each condition embedding as its response score. As shown in Fig.\ref{response_scores}, we can see that the response score of condition embeddings is dynamic over diffusion steps and condition types. For example, $E_{B}$ has a stronger response at the earlier diffusion steps than at the later steps.}
 
	\textbf{\textit{Effects of Training Loss.}} In this section, we explore the effects of $\mathcal{L}_{\mathcal{V}}$ loss and investigate its hyper-parameter configuration of loss weight $\lambda_{kl}$. As shown in Tab.\ref{abl_loss}, we first report the human motion prediction with the $\mathcal{L}_{mr}$-only training loss (top row). Then, by introducing additional $\mathcal{L}_{kl}$ training loss into $\mathcal{L}_{\mathcal{V}}$, we see that it brings clear human motion prediction performance gains. Then, we further tune the loss weight of $\mathcal{L}_{kl}$ from $10^{-1}$ to $10^{-5}$ to investigate its optimal configuration choice. Analyzing the results shown in Tab.\ref{abl_loss}, we can find that MCLD is insensitive to the choice of $\mathcal{L}_{kl}$. Since $\mathcal{L}_{kl}=10^{-4}$ brings a slightly better performance, we thus choose it as our default configuration in the final model deployment.

	\begin{table}[t]
	\caption{Performance comparison between different configurations of $\mathcal{L}_{\mathcal{V}}$. The best performances are indicated with bold.}
		\centering
		\resizebox{0.48\textwidth}{!}{
		\begin{tabular}{c|c|cccc}
			\toprule
			\multicolumn{2}{c|}{$\mathcal{L}_{\mathcal{V}}$ Configuration}        & \multicolumn{4}{c}{3D Pose Error $\downarrow$} \\ \hline
			Loss Term & \multicolumn{1}{c|}{Loss Weight} & 0.5s     & 1.0s    & 1.5s    & 2.0s    \\ \hline
		   	$\mathcal{L}_{mr}$-only &          $\lambda_{mr}=1$   & 53 &  72 &  85  & 97 \\ \hline   
		   	\multirow{5}{*}{\begin{tabular}[c]{@{}c@{}}$\mathcal{L}_{mr}$ \\ \&\\ $\mathcal{L}_{kl}$\end{tabular}}& $\lambda_{mr}=1$, $\lambda_{kl}=10^{-1}$ & 44 &  63 &  69  & 82        \\
		   	& $\lambda_{mr}=1$, $\lambda_{kl}=10^{-2}$    &    43      &    61     &     67    &   82      \\
		   	& $\lambda_{mr}=1$, $\lambda_{kl}=10^{-3}$    &    43      &    61     &  \textbf{65}    &  \textbf{81}      \\
		   	& $\lambda_{mr}=1$, $\lambda_{kl}=10^{-4}$    & \textbf{42} & \textbf{60} & \textbf{65} & \textbf{81}  \\
		   	& $\lambda_{mr}=1$, $\lambda_{kl}=10^{-5}$    &    43      &  \textbf{60} & \textbf{65} &  82      \\ \bottomrule		   	  
		\end{tabular}
	}
	\label{abl_loss}
	\end{table}

\section{Limitation and Future Work}
 \revise{In this section, we briefly analyze the underlying limitation of our method to inspire its future development. We consider the current MCLD as a static model as its some hyper-parameter configurations are fixed over all input samples, such as the layer number of its multi-attention encoder, the number of its denoising steps, the down-sample rate of its 3D scene point cloud, etc. In the future, we would make efforts to develop MCLD into a dynamic one that can adaptively infer these configurations on each input sample, significantly improving real-time human motion prediction.} 
	\section{Conclusion}
	In this paper, instead of planning future human motion in a “dark” room, we propose a multi-condition latent diffusion network that reformulates the human motion prediction task as a multi-condition joint inference problem based on the historical 3D body motion and the current 3D scene context conditions, significantly improving future human motion prediction. To reduce the redundancy of initial 3D scene input, we propose a key region proposal module to adaptively infer a localized interaction-related region for each human motion sample. Furthermore, we develop a powerful multi-attention encoder to extract multiple dependencies within and between 3D body joints and scene points, including body dynamics, scene geometries, and body-scene interactions. By coupling all these proposed schemes, we develop a powerful 3D human motion prediction system that outperforms state-of-the-art methods on both realistic and diverse predictions.  
	
\bibliographystyle{IEEEtran}
\bibliography{IEEEabrv,ref}


\end{document}